\documentclass{article}

 \usepackage[preprint]{neurips_2025}

\usepackage[utf8]{inputenc} 
\usepackage[T1]{fontenc}    
\usepackage{hyperref}       
\usepackage{url}            
\usepackage{booktabs}       
\usepackage{amsfonts}       
\usepackage{nicefrac}       
\usepackage{microtype}      

\usepackage{multirow}
\usepackage{multicol}
\usepackage{makecell}
\usepackage{threeparttable}
\usepackage{pifont}
\usepackage{enumitem}
\usepackage{xcolor}
\usepackage{graphicx}
\usepackage{subcaption}
\usepackage{amsmath}
\usepackage{wrapfig}

\newcommand{\sysname}{Face4FairShifts}
\newcommand{\utkfair}{UTK-FairFace}

\title{\sysname{}: A Large Image Benchmark for Fairness and Robust Learning across Visual Domains}

\author{%
  Yumeng Lin$^1$\thanks{Equal Contribution.} $\quad$ Dong Li$^{2\ast}$ $\quad$ Xintao Wu$^3$ $\quad$ Minglai Shao$^1$ $\quad$ Xujiang Zhao$^4$ \\ \textbf{Zhong Chen}$^5$ $\quad$ \textbf{Chen Zhao}$^2$\\
  $^1$School of New Media and Communication, Tianjin University\\
  $^2$Department of Computer Science, Baylor University\\
  $^3$Electrical Engineering and Computer Science Department, University of Arkansas\\
  $^4$NEC Laboratories America\\
  $^5$School of Computing, Southern Illinois University\\
  \texttt{
  \{lym619,shaoml\}@tju.edu.cn, \{dong\_li1,chen\_zhao\}@baylor.edu
  } \\
  \texttt{
  xintaowu@uark.edu, xuzhao@nec-labs.com, zhong.chen@cs.siu.edu
  }
}

\begin{document}

\maketitle

\begin{abstract}
  Ensuring fairness and robustness in machine learning models remains a challenge, particularly under domain shifts. We present \textsc{\sysname{}}, a large-scale facial image benchmark designed to systematically evaluate fairness-aware learning and domain generalization. The dataset includes 100K images across four visually distinct domains with 42 annotations within 15 attributes covering demographic and facial features. Through extensive experiments, we analyze model performance under distribution shifts and identify significant gaps. Our findings emphasize the limitations of existing related datasets and the need for more effective fairness-aware domain adaptation techniques. \textsc{\sysname{}} provides a comprehensive testbed for advancing equitable and reliable AI systems. The dataset is available online at \url{https://meviuslab.github.io/Face4FairShifts/}.

\end{abstract}

\section{Introduction}
\label{sec:introduction}
Distribution shifts between source and target domains present a fundamental challenge in machine learning, affecting model generalization in real-world applications. For example, a facial recognition system trained on high-resolution studio portraits may perform poorly when applied to low-quality surveillance footage, leading to accuracy degradation and biased predictions. Such shifts in visual characteristics highlight the need for robust adaptation techniques to ensure model reliability.
Broadly, as shown in Table \ref{tab:shifts}, distribution shifts can be categorized into five key types: 
(1) covariate shifts \cite{shimodaira2000improving}, 
(2) semantic shifts \cite{wang2003mining}, 
(3) demographic shifts \cite{giguere2022fairness}, 
where the marginal distribution of input features, class labels, and sensitive attributes (race and gender) changes across domains, respectively;
(4) concept shifts \cite{conceptshift}, where the relationship between input features and labels changes; 
and (5) correlation shifts \cite{roh2023improving,fedora}, where spurious correlations between sensitive attributes and class labels vary across domains. 
These shifts, whether occurring independently or in combination, can significantly impact both predictive performance and fairness, making them critical considerations in real-world deployment.

Recently, fairness-aware out-of-distribution (OOD) generalization (FairOG) \cite{fedora,fatdm,flair,utk-fairface,fvae,eiil}, has emerged as an important research area, addressing the joint effects of covariate and correlation shifts. The FairOG setting assumes access to training data from multiple source domains and aims to learn a fair classifier that generalizes to an unseen and inaccessible target domain while maintaining both predictive accuracy and fairness. However, studying FairOG in practice is challenging due to the
\begin{wraptable}{r}{0.55\textwidth}
    \centering
    \setlength\tabcolsep{2.5pt}
    \tiny
    \caption{Different Types of Distribution Shifts \cite{shao2024supervised,fedora}. Let $X,Y,Z$ represent variables of data features, class labels, and sensitive attributes, respectively, with superscripts $s,t$ indicating source and target domains.}
    \begin{tabular}{lll}
        \toprule
         \textbf{Type of Shifts} & \textbf{Notations} & \textbf{Descriptions of Shifts}\\
        \midrule
        Covariate Shifts \cite{shimodaira2000improving} & $\mathbb{P}_{X}^s\neq\mathbb{P}_{X}^t$ & Marginal distribution shifts on $X$\\
        Semantic Shifts \cite{wang2003mining} & $\mathbb{P}_{Y}^s\neq\mathbb{P}_{Y}^t$ & Marginal distribution shifts on $Y$\\
        Demographic Shifts \cite{giguere2022fairness} & $\mathbb{P}_{Z}^s\neq\mathbb{P}_{Z}^t$ & Marginal distribution shifts on $Z$\\
        Concept Shifts \cite{conceptshift} & $\mathbb{P}_{Y|X}^s\neq\mathbb{P}_{Y|X}^t$ & Conditional distribution shifts on $Y|X$\\
        Correlation Shift \cite{roh2023improving} & $\mathbb{P}_{Y,Z}^s\neq\mathbb{P}_{Y,Z}^t$ & Joint distribution shifts on $(Y,Z)$\\
        \bottomrule
    \end{tabular}
    \label{tab:shifts}
    \vspace{-3mm}
\end{wraptable}
lack of suitable benchmark datasets. Existing datasets \cite{utk-fairface,fairface} are often repurposed by artificially partitioning a single dataset into multiple domains \cite{fedora} or by defining sensitive attributes post hoc \cite{eiil,fatdm}. These adaptations frequently fail to capture meaningful domain-specific covariate shifts or strong correlations between sensitive attributes and class labels, limiting their effectiveness in evaluating FairOG methods.
A detailed discussion of related benchmark datasets is provided in Appendix~\ref{sec:relatedwork}.
To enable rigorous research in FairOG, there is a pressing need for a dataset that naturally exhibits covariate shifts across multiple domains while simultaneously demonstrating correlation shifts between sensitive attributes and class labels. Such a dataset would provide a more realistic and robust testbed for developing and evaluating fairness-aware generalization techniques under real-world distribution shifts.

In this paper, we introduce, \textsc{\sysname{}}, a large-scale facial image benchmark designed to study fairness-aware domain generalization and robustness under distribution shifts.
\textsc{\sysname{}} consists of 100,000 facial images spanning four visually distinct domains, Photo, Art, Cartoon, and Sketch, each exhibiting meaningful covariate shifts.
The dataset is constructed using a combination of existing datasets and web-crawled images, ensuring diverse and representative domain characteristics.
Each image in \textsc{\sysname{}} is annotated with 42 annotations across 15 attributes, capturing various demographic and facial characteristics.
To ensure high-quality annotations, we employed 66 human annotators and a dedicated 5-person quality control team to refine and validate the labels.
We conduct extensive experiments to evaluate \textsc{\sysname{}} across four key machine learning research areas: fairness learning, OOD generalization, OOD detection, and FairOG. 
In each area, we compare \textsc{\sysname{}} with existing datasets that have been adapted for the corresponding problem setting and evaluate it against state-of-the-art methods to ensure a comprehensive benchmark.
Our results show that \textsc{\sysname{}} presents significant challenges to current baseline models, highlighting the difficulties of achieving fairness and robustness under realistic domain shifts.
By providing a carefully curated benchmark with meaningful distribution shifts, \textsc{\sysname{}} serves as a valuable resource for advancing research in fair and robust machine learning.
The main contribution of this paper is summarized:
\begin{itemize}[leftmargin=*]
    \item We introduce \textsc{\sysname{}}, a novel large-scale facial dataset specifically designed for studying fairness and robust machine learning. The dataset comprises 100,000 images spanning four distinct visual domains and includes 42 annotations across 15 attributes, capturing a diverse range of facial features.
    
    \item Images in \textsc{\sysname{}} are sourced from existing datasets and crawled from search engines. 
    We recruit 66 paid annotators to perform over 12.5 million annotation instances and a 5-person team to refine and validate the labels, making the annotation process highly labor-intensive, and requiring substantial effort and resources.
    
    \item Extensive empirical studies on various state-of-the-art baseline methods across four research areas, including fairness learning, OOD generalization, OOD detection, and fairness-aware OOD generalization, demonstrate that \textsc{\sysname{}} presents greater challenges compared to existing datasets while exhibiting more stable performance across baseline methods.
    These findings establish \textsc{\sysname{}} as a valuable resource for developing and evaluating fair and robust machine learning techniques.
\end{itemize}



\section{The \textsc{\sysname{}} Dataset}
\label{sec:dataset}

\subsection{Data Collection}
Our \textsc{\sysname{}} dataset collection process consists of two steps. 
First, we gather publicly available datasets and apply filtering to ensure data quality and relevance. 
Second, if a domain lacks sufficient data after filtering, we supplement it by crawling images from search engines. 
Specifically, for the four domains (Photo, Art, Cartoon, and Sketch) in \textsc{\sysname{}}, all images in the Photo and Cartoon domains come from existing datasets, while the majority in the Art and Sketch domains are obtained through web crawling due to the limited availability of public datasets. 
This approach ensures a diverse and comprehensive dataset while maintaining consistency across domains.

\begin{figure*}[t]
    \centering
    \includegraphics[width=\linewidth]{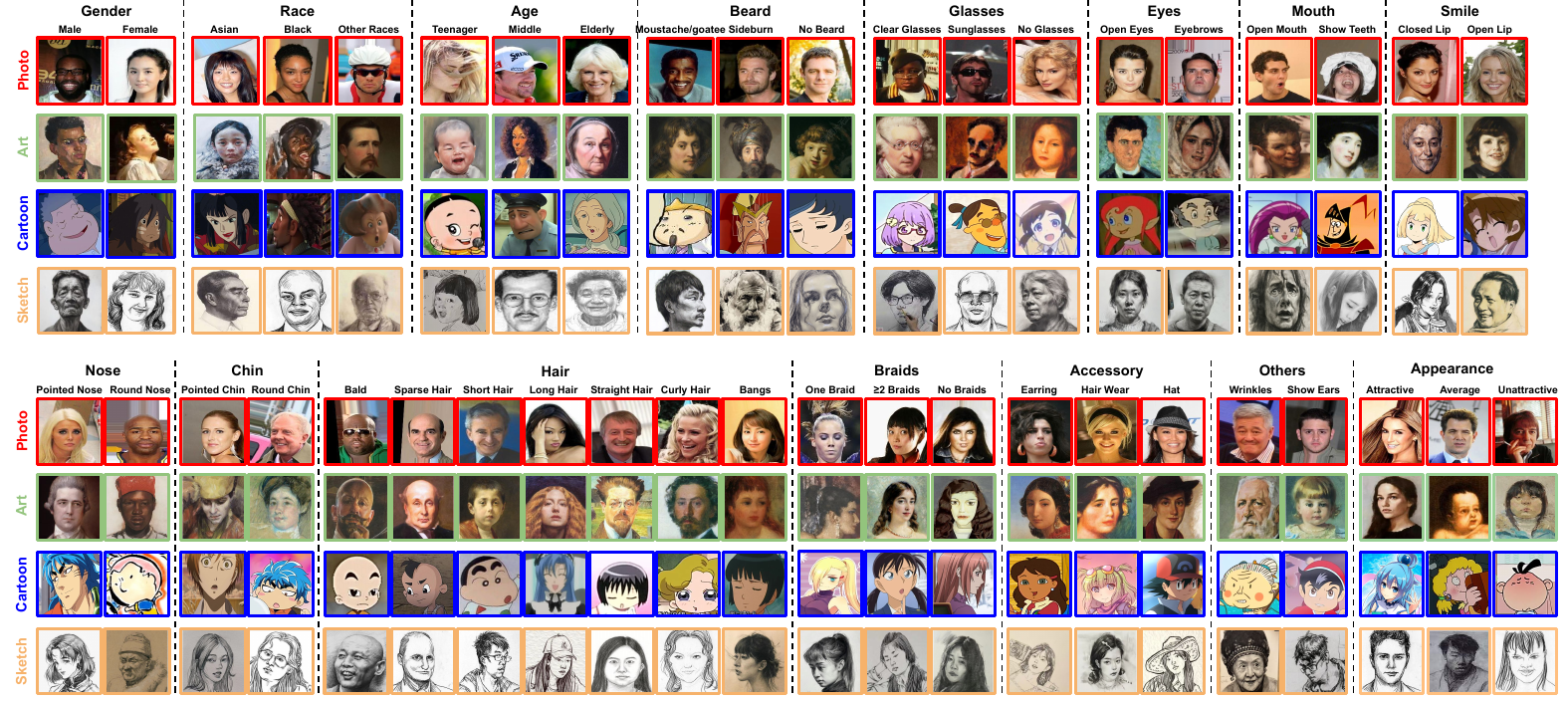}
    \vspace{-5mm}
    \caption{An example of the \textsc{\sysname{}} dataset, showcasing four domains (Photo, Art, Cartoon, and Sketch) with 42 annotations across 15 attributes. 
    }
    \label{fig:dataexp}
    \vspace{-3mm}
\end{figure*}

\textbf{Collection from Existing Facial Datasets.}
We systematically collect existing datasets and carefully select the most suitable ones to construct each domain of \textsc{\sysname{}}. 
These datasets include not only facial image datasets but also full-body and half-body image datasets, which require cropping to align with our requirements.
We found that no existing facial dataset spans all four designed domains, highlighting the significance and necessity of the \textsc{\sysname{}} dataset.

\begin{itemize}[leftmargin=*]
    \item \textbf{Photo Domain.}
    \textsc{CelebA} \cite{CelebA} is a large-scale celebrity face dataset with rich annotations, featuring a diverse range of images across different races, age groups, and other characteristics. Its diversity makes it well-suited for representing face images in the Photo domain. Based on this, we select 30,000 images from \textsc{CelebA}. Additionally, the detailed annotations in \textsc{CelebA} serve as valuable references for \textsc{\sysname{}}.

    \item \textbf{Art Domain.}
    \textsc{MetFaces} \cite{karras2020training} is a face dataset extracted from artwork, comprising 1,336 high-quality images obtained via the \textit{Metropolitan Museum of Art Collection Application Programming Interface} \cite{API}. To maintain a clear distinction between the Art and Sketch domains, we exclude black-and-white images from this dataset. The \textsc{WikiART} \cite{artgan2018} dataset contains 6,095 portrait face images scraped using the \textit{WikiART crawler} \cite{artgan2018}, covering a diverse range of artistic styles, with black-and-white images similarly excluded. The \textsc{Human-Art} \cite{ju2023human} dataset, proposed by the \textit{Chinese University of Hong Kong}, is a multifunctional dataset featuring images across various categories, with 2,000 images per category. For the Art domain in \textsc{\sysname{}}, we include images from the watercolor and oil painting categories. 

    \item \textbf{Cartoon Domain.}
    \textsc{IIIT-CFW} \cite{mishra2016iiit} is a cartoon face dataset containing 8,928 images of 100 world-renowned individuals from various professions. Due to variations in image size, with some being small and low-resolution, we only utilize a carefully selected subset of high-quality images. \textsc{iCartoonFace} \cite{zheng2020cartoon} is a large-scale face recognition dataset comprising 389,678 images of 5,013 cartoon characters collected from public websites and online videos. The majority of the images in the Cartoon domain of \textsc{\sysname{}} come from \textsc{iCartoonFace}.

    \item \textbf{Sketch Domain.}
    The \textsc{CUFSF} \cite{zhang2011coupled} dataset includes face images with lighting variation and one corresponding sketch for each of its 1,194 individuals. We include these 1,194 sketches in the Sketch domain of \textsc{\sysname{}}. \textsc{FS2K} \cite{fan2022facial} is a high-quality facial sketch dataset consisting of 2,104 sketch images created by professional artists in three distinct styles. Overall, the available data for the Sketch domain remains limited.
    
\end{itemize}

\textbf{Image Crawling via Search Engines.}
After collecting existing datasets, the total number of images remained significantly below our target (100K). To address this, we developed a Python-based web crawler to gather additional images from multiple search engines, including \textit{Google} and \textit{Bing}. The images were primarily sourced from various platforms such as \textit{Xiaohongshu}, \textit{YouTube}, \textit{Douyin}, \textit{Saatchi Art}, \textit{Etsy}, \textit{eBay}, \textit{Blibli}, and \textit{Sina}.

\subsection{Data Processing}
\begin{figure}[t]
    \centering
        \subfloat{\includegraphics[width=0.35\linewidth]{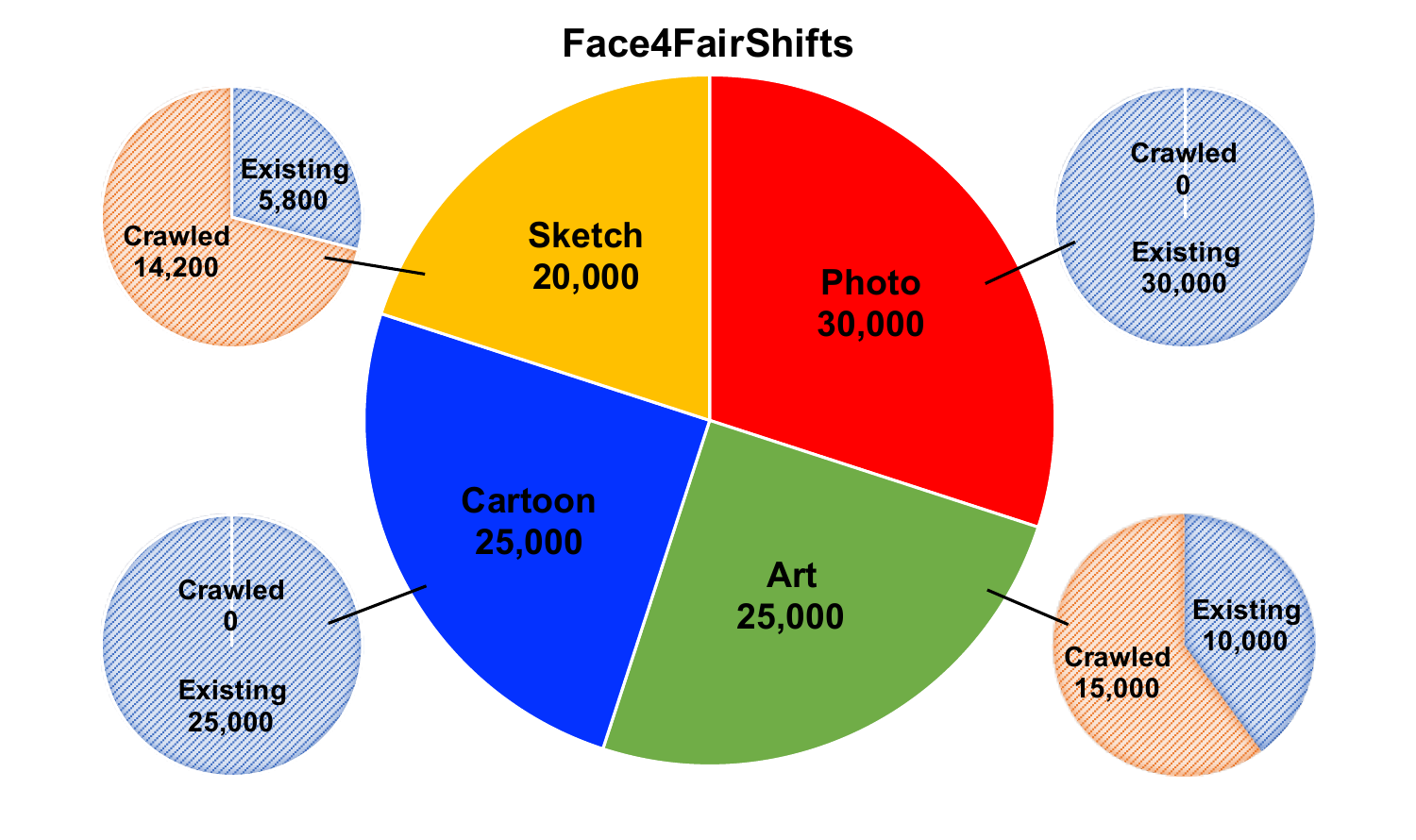}}
        \subfloat{\includegraphics[width=0.65\linewidth]{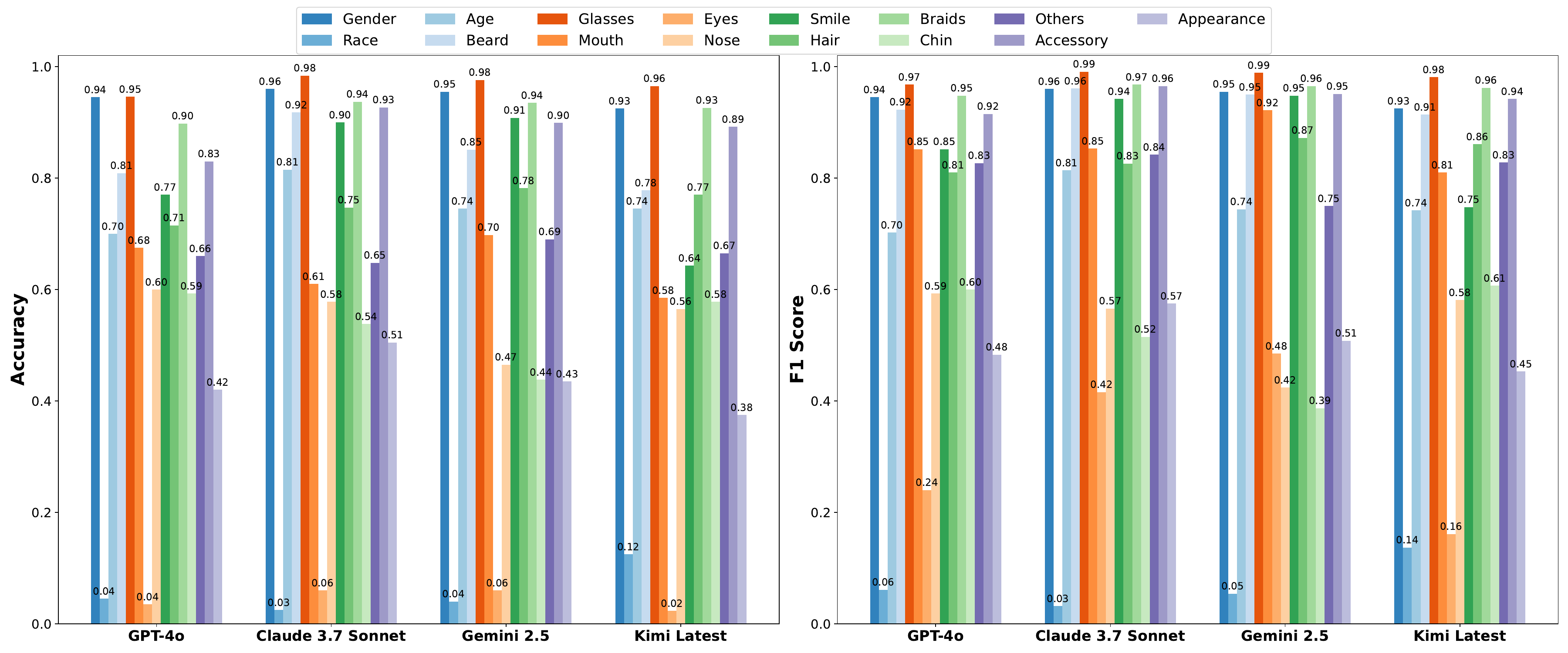}}
    \caption{\textbf{(Left)} An overview of the \textsc{\sysname{}} dataset after data processing, comprising images collected from existing datasets and supplemented by web-crawled data across all domains.
    \textbf{(Right)} Comparison of four large language models (GPT-4o, Claude 3.7 Sonnet, Gemini 2.5, and Kimi Latest) on 15 facial attribute categories in terms of accuracy and F1-score, using human annotations as ground truth.
    }
    \label{fig:datacollection}
    \vspace{-5mm}
\end{figure}

In the collected data, particularly those obtained through search engines, it is inevitable to encounter images that either do not contain any individuals or feature multiple people. 
For person detection, we utilize the YOLOv5 object detection tool~\cite{glenn_jocher_2020_3983579}, which allows us to systematically process and filter the collected images. It removes images where no person is detected and segments each detected individual into a separate image, ensuring that each output image contains only one person.
Specifically, for each crawled image, we focus exclusively on detecting the "person" class, setting a confidence threshold of 0.7. If the confidence score for detecting a person in an image falls below this threshold, the object is disregarded, while those meeting the criterion are retained. However, this process introduces a new challenge: some of the segmented individuals are too small, making them unsuitable for our dataset. To mitigate this issue, we manually determine a minimum acceptable image size as a reference. We then iterate through all cropped images and discard those that fall below this size threshold, ensuring sufficient clarity and quality.
Additionally, since the cropped images of detected individuals may include half-body or full-body portraits, and automated face cropping may not be reliable for achieving consistent image framing, manual cropping using \textit{Photoshop} is required. Although labor-intensive, this final step was necessary to ensure high-quality, standardized image framing across the dataset, maintaining consistency for subsequent analysis.
Figure~\ref{fig:datacollection} (Left) illustrates the dataset statistics following the data processing stage.

\subsection{Data Annotation}

\begin{figure*}[t]
    \centering
    \includegraphics[width=\linewidth]{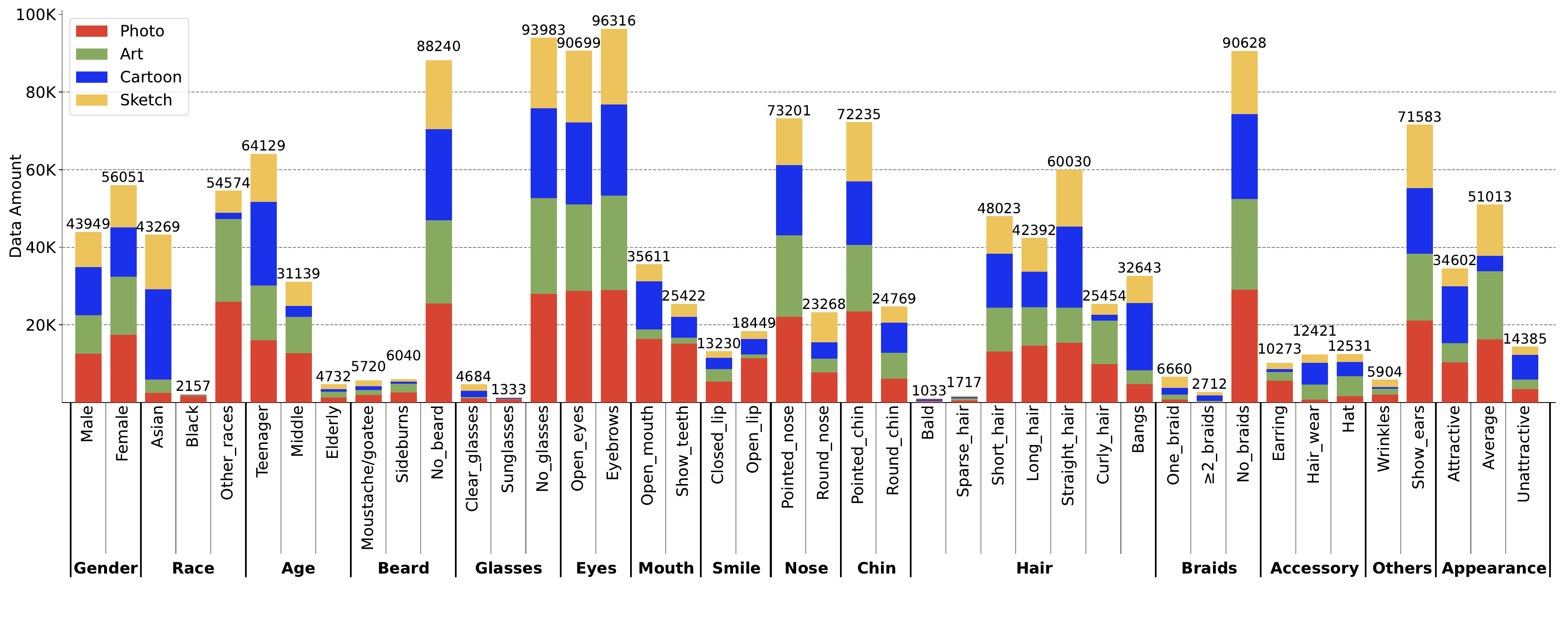}
    \vspace{-8mm}
    \caption{
    Statistical overview of annotations in the \sysname{} dataset. The number at the top of each column indicates the total number of images for each annotation.
    }
    \label{fig:annostatistics}
    \vspace{-3mm}
\end{figure*}

Each processed face image in \textsc{\sysname{}} is annotated with 42 binary labels across 15
attributes, covering a diverse range of features. Our annotation scope is extensive, encompassing sensitive attributes related to fairness, such as gender, race, and age, as well as detailed facial attributes including beard, glasses, eyes, mouth, smile, nose, chin, hair, braids, accessories, and others. To enhance the dataset’s versatility and enable fine-grained analysis, each facial attribute is further divided into multiple specific annotations, such as male, female, Asian, black, and so forth.
All annotations in \textsc{\sysname{}} are manually labeled to ensure high quality. 
An example in Figure \ref{fig:annoexp} of the Appendix illustrates the observed and unobserved labels for a given image in the Art domain.
We recruited 66 paid annotators and provided them with comprehensive training sessions to ensure a thorough understanding of the annotation guidelines. To maintain consistency and accuracy, each image is labeled by at least three annotators, and the final annotation was determined using a majority voting mechanism, where the label agreed upon by the majority was assigned as the final decision.
To further enhance data quality, a five-person quality control team conducted random sample reviews to verify annotation accuracy. The total number of annotation instances in the \textsc{\sysname{}} dataset exceeds 12,600,000, making this a highly labor-intensive process that requires substantial effort and resources.
Figure \ref{fig:dataexp} showcases example facial images, and Figure \ref{fig:annostatistics} summarizes the distribution of 42 annotations across the four domains.

Furthermore, we conduct a comparative analysis between human annotations and those produced by large language models (LLMs). We randomly selected 500 images from each of the four domains, resulting in a total of 2000 images. These images were then input into four representative LLMs: GPT-4o, Claude 3.7 Sonnet, Gemini 2.5, and Kimi Latest.  A carefully designed prompt was constructed to guide the models in performing visual attribute annotation. Figure \ref{fig:datacollection} (Right) presents the performance comparison of the four LLMs in terms of accuracy and F1-score across all 15 facial attribute categories, demonstrating the consistency and reliability of human annotations as a benchmark.
More details of data annotations refer to Appendix \ref{sec:app_dataannotation}.

\subsection{Fairness Statistics}

\begin{figure*}[t]
    \begin{subfigure}{\textwidth}
        \centering
        \includegraphics[width=\linewidth]{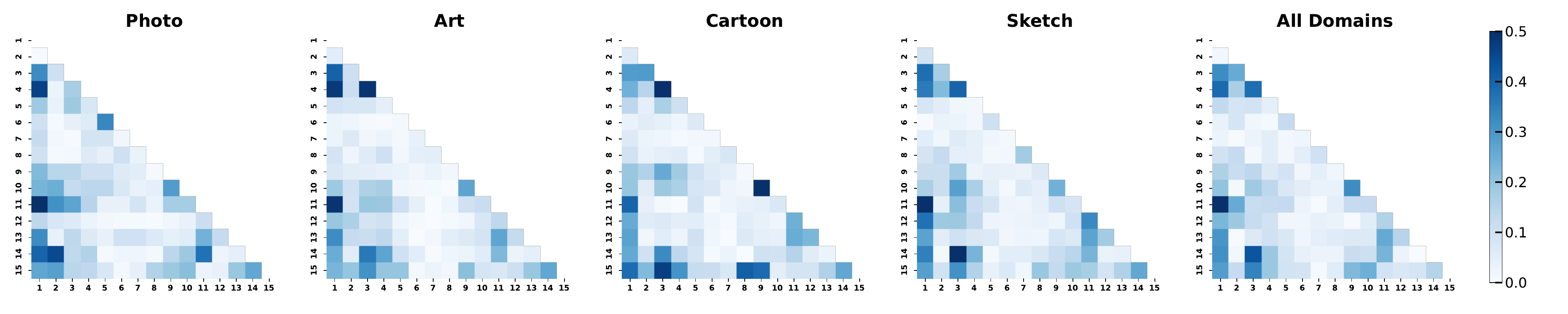}
    \end{subfigure}
    \hfill
    \begin{subfigure}{\textwidth}
        \centering
        \includegraphics[width=\linewidth]{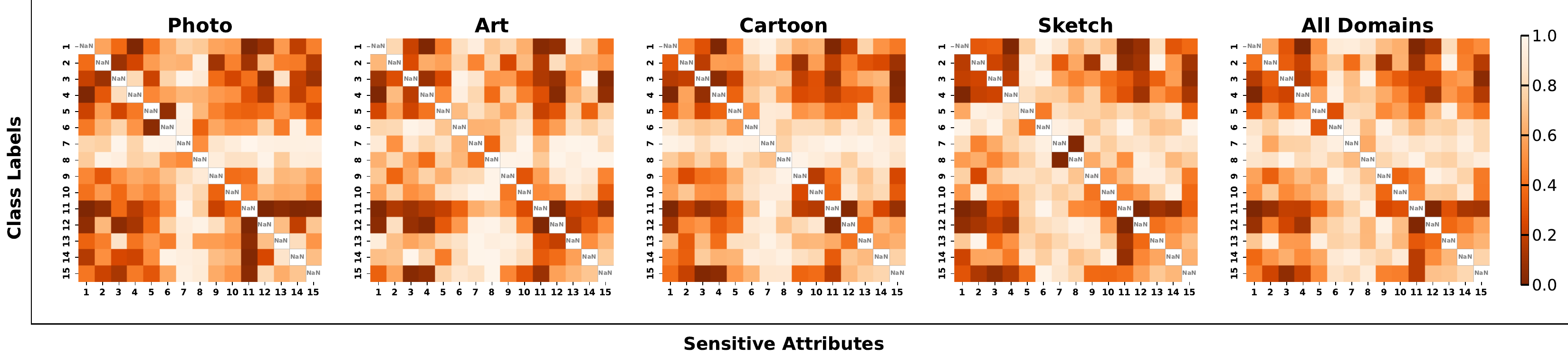}
    \end{subfigure}
    \vspace{-5mm}
    \caption{
    Visualization of fairness statistics in \textsc{\sysname{}} using heatmaps across all domains. Numbers on the $x$-axis and $y$-axis represent attributes Gender (1), Race (2), Age (3), Beard (4), Glasses (5), Eyes (6), Mouth (7), Smile (8), Nose (9), Chin (10), Hair (11), Braids (12), Accessory (13), Others (14), and Appearance (15).
    \textbf{(Upper)} Each cell represents the absolute correlation coefficient between two attributes, indicating their statistical dependence.
    \textbf{(Lower)} Each cell represents the disparate impact (DI) of a sensitive attribute on a class label, measuring fairness disparities.
    Darker cells in all heatmaps indicate stronger dependence, highlighting greater bias or disparities in \textsc{\sysname{}}.
    }
    \label{fig:heatmap}
    \vspace{-5mm}
\end{figure*}

To evaluate the fairness characteristics of \textsc{\sysname{}}, we perform a statistical analysis across all four domains. This analysis aims to quantify fairness-related biases and disparities both within each domain and the dataset as a whole, providing insights into how fairness varies across different combinations of sensitive attributes and labels.
Specifically, we examine the absolute correlation between all 14 attributes using heatmaps, which visually represent how these attributes interact within each combination.
Furthermore, we compute disparate impacts (DI) to assess fairness disparities.
By measuring the degree to which positive class labels ($Y=1$) are influenced by sensitive attributes ($Z$), we quantify whether certain sensitive groups are disproportionately advantaged or disadvantaged. 
For this analysis, we discretize these attributes into binary groups, defining race as Black and non-Black, age as elderly and non-elderly, beard as has-beard and no-bread, glasses as has-glasses and no-glasses,  hairs as bald and no-bald, braids as has-braids and no-braids, and accessory as has-hat and no-hat. The disparate impact is then calculated as:
{
\small
\begin{align*}
    \text{DI} = k, \text{if DI}\leq1; \text{DI}=1/k, \text{otherwise},\:\: \text{where } k=\frac{\mathbb{P}(Y=1|Z=-1)}{\mathbb{P}(Y=1|Z=1)} 
\end{align*}
}
Following the "80\%-rule" \cite{biddle2017adverse}, DI values greater than or equal to 0.8 indicate that the data meets fairness criteria.
Figure \ref{fig:heatmap} highlights substantial correlation shifts between sensitive attributes and class labels across domains, demonstrating the challenges of fair model adaptation when data distributions shift. These findings suggest that models trained on one domain may exhibit unfair biases when generalized to another domain, emphasizing the need for fairness-aware domain adaptation techniques.
As a result, \textsc{\sysname{}} serves as a valuable benchmark for fairness-aware domain generalization, providing a robust testbed for studying bias mitigation techniques under both covariate and correlation shifts. 
By exposing fairness disparities in real-world facial datasets across multiple visual styles, \textsc{\sysname{}} facilitates the development of more equitable, unbiased, and robust machine learning models.


\section{Experiments}
\label{sec:experiments}
\begin{figure}[t]
    \begin{subfigure}{\textwidth}
        \centering
        \subfloat{\includegraphics[width=0.248\linewidth]{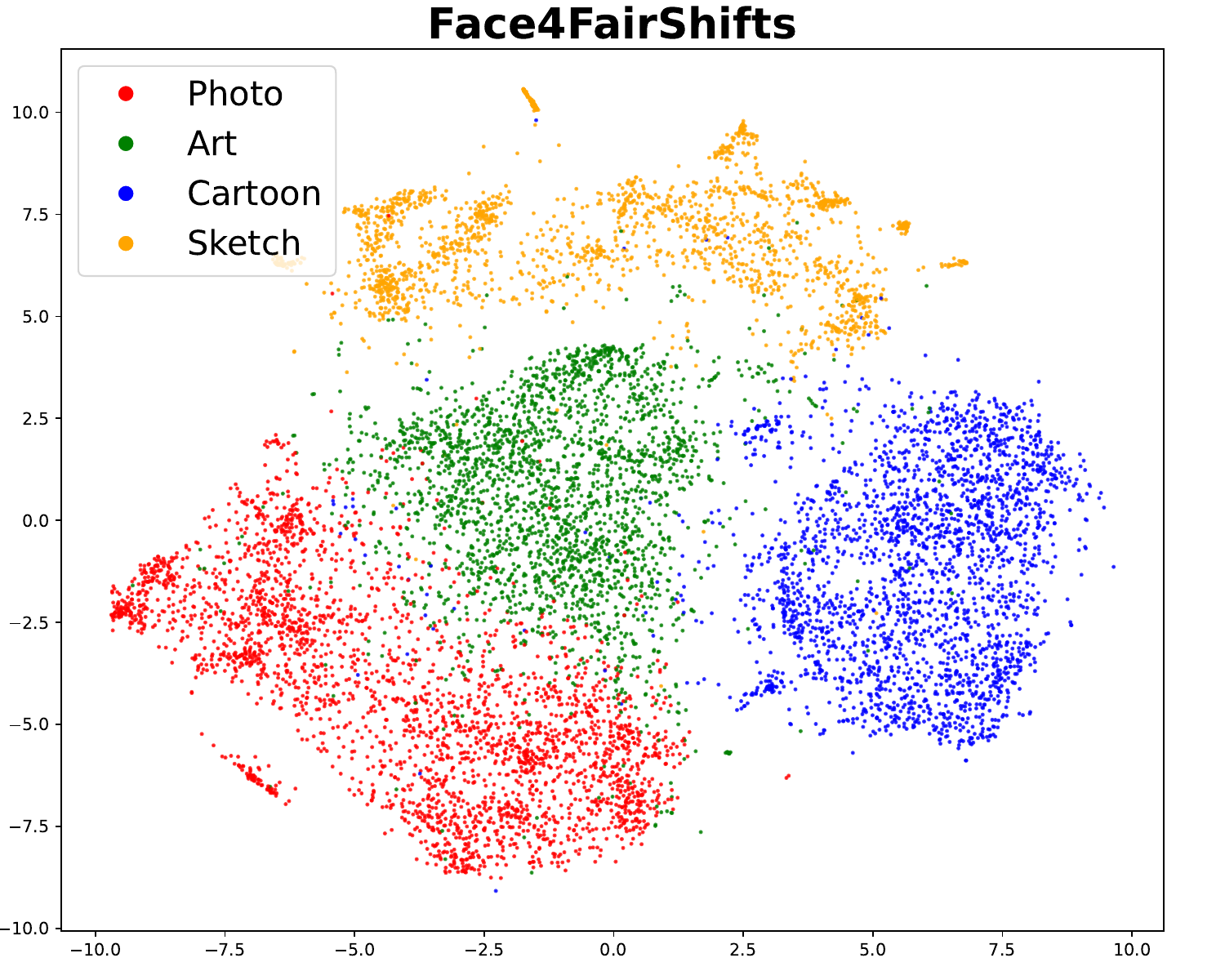}}
        \subfloat{\includegraphics[width=0.248\linewidth]{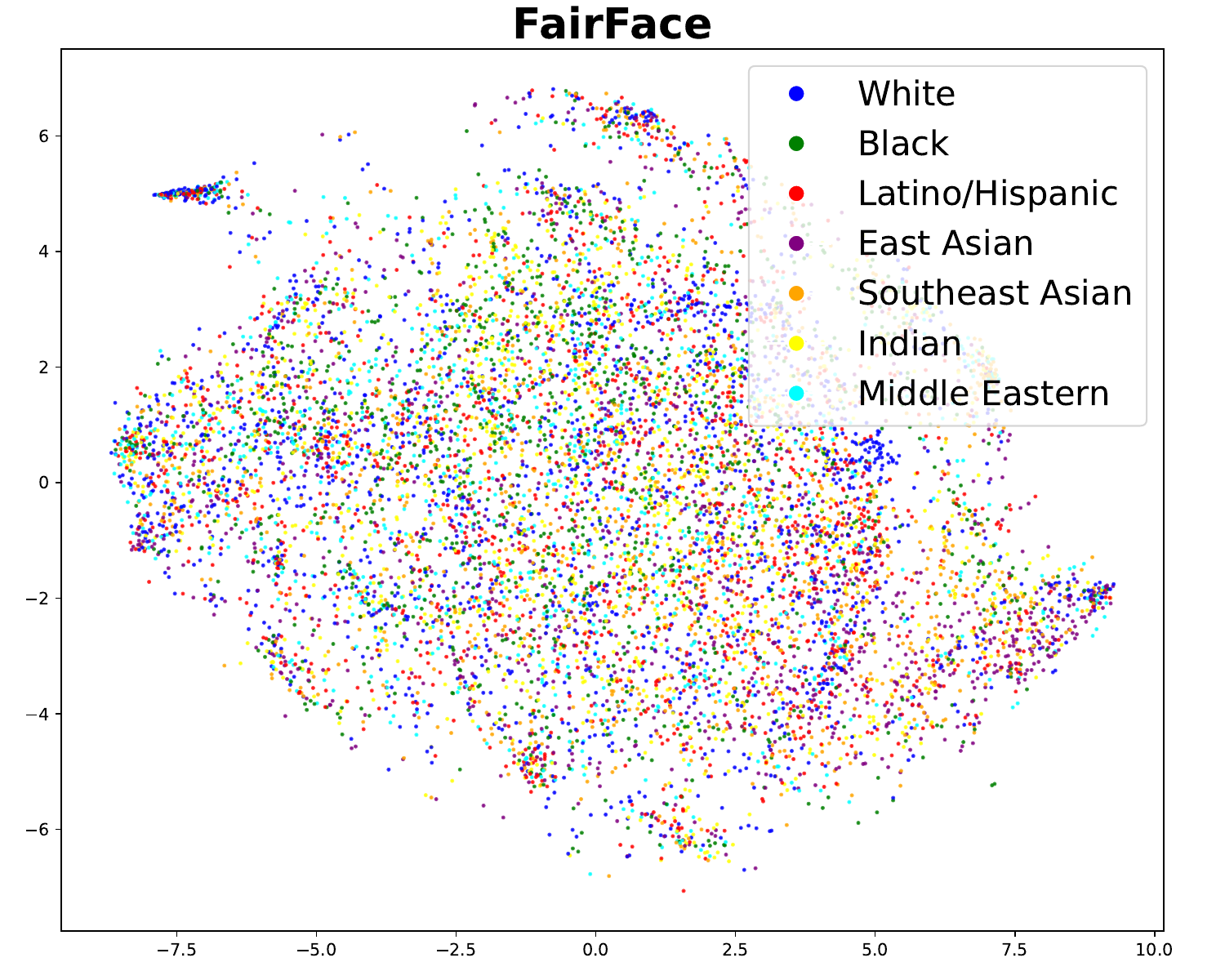}}
        \subfloat{\includegraphics[width=0.248\linewidth]{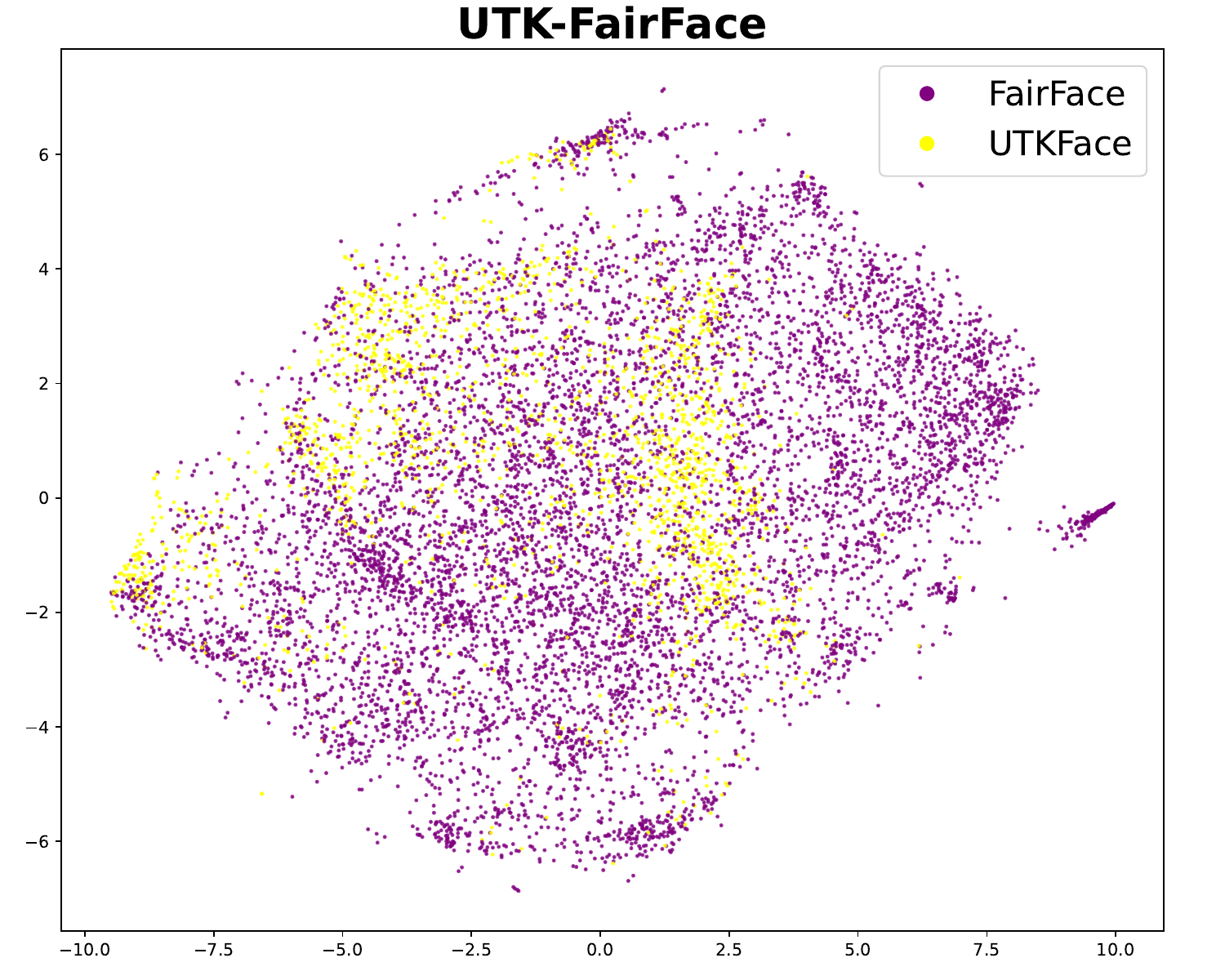}}
        \subfloat{\includegraphics[width=0.210\linewidth]{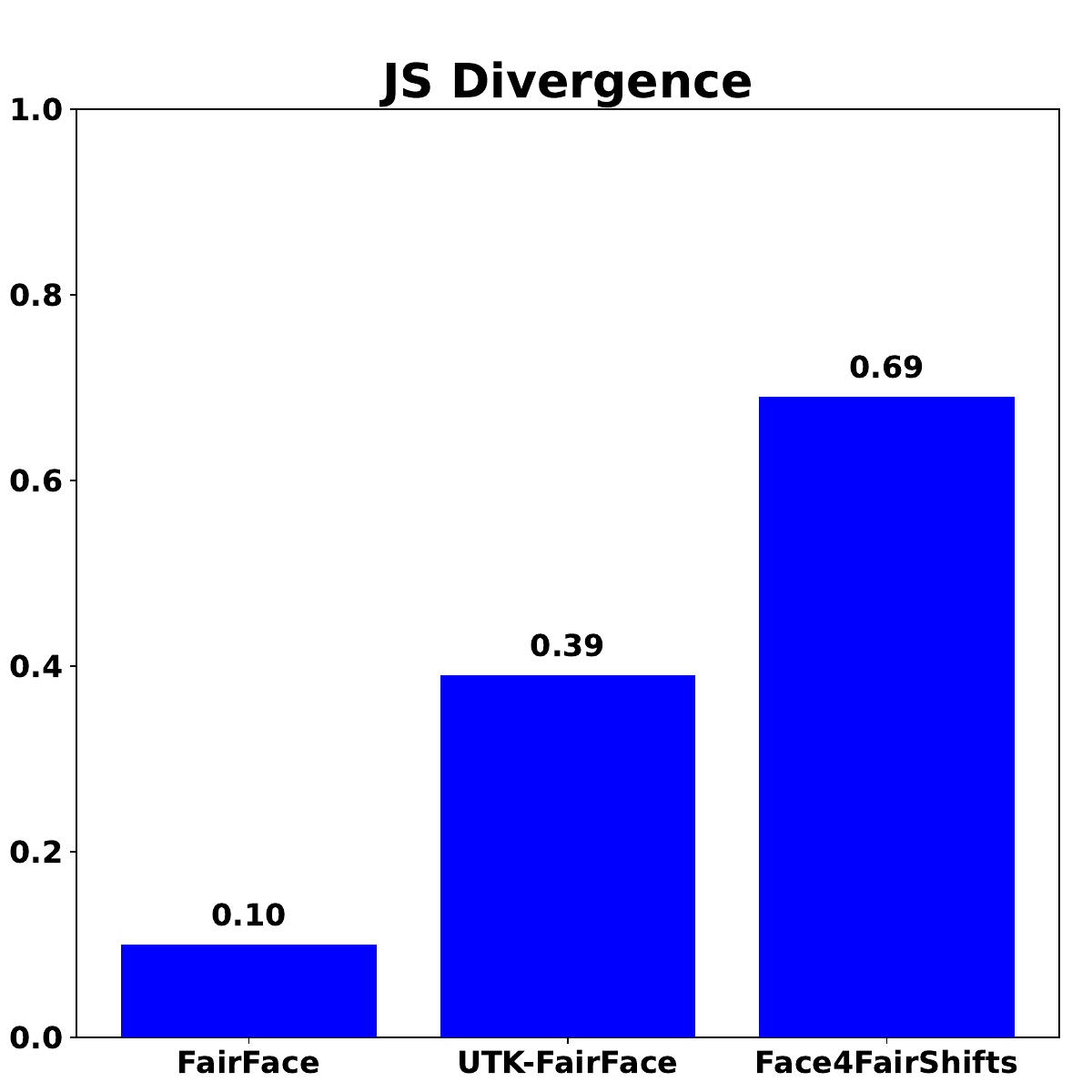}}
    \end{subfigure}
    \hfill
    \begin{subfigure}{\textwidth}
        \centering
        \subfloat{\includegraphics[width=0.248\linewidth]{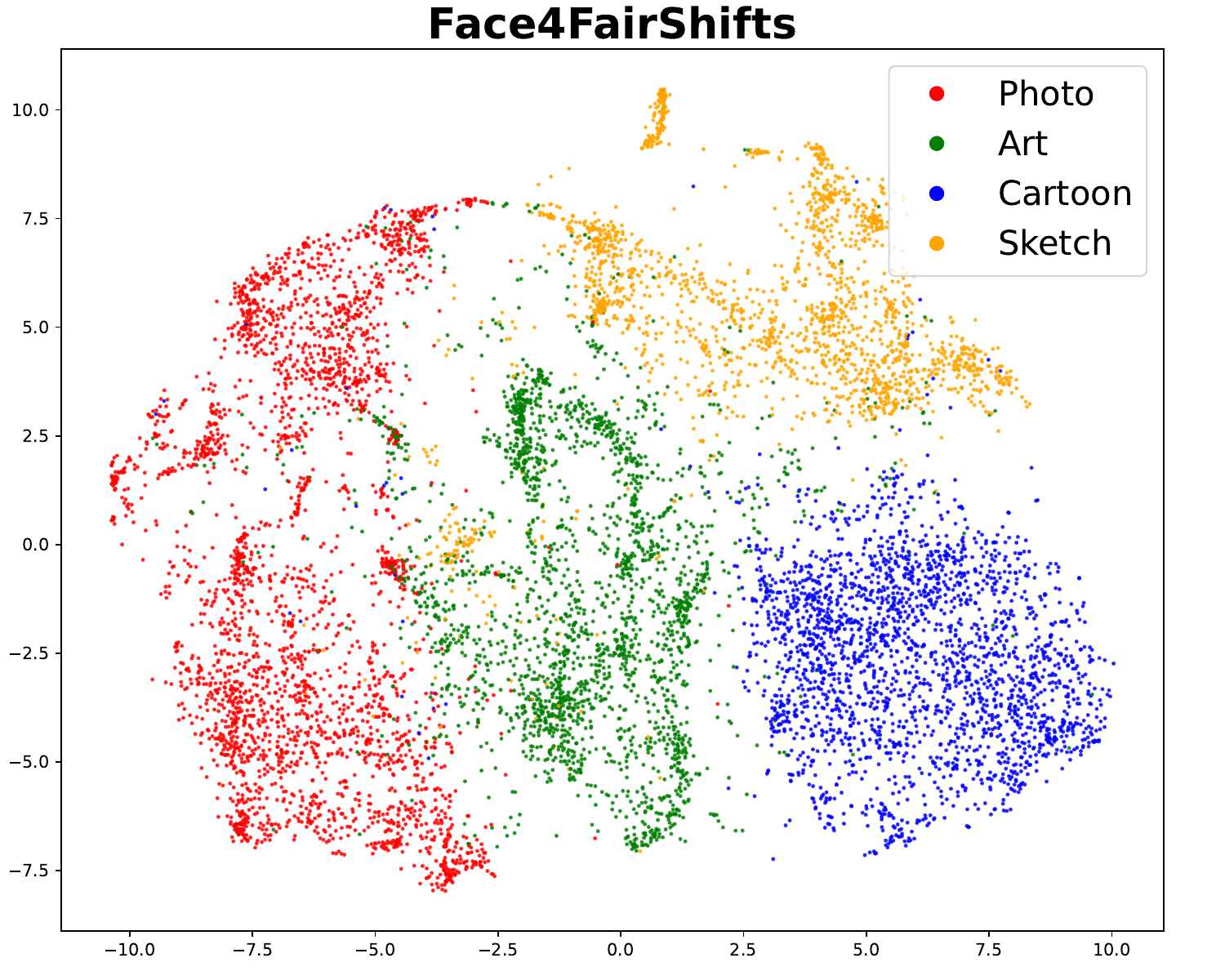}}
        \subfloat{\includegraphics[width=0.248\linewidth]{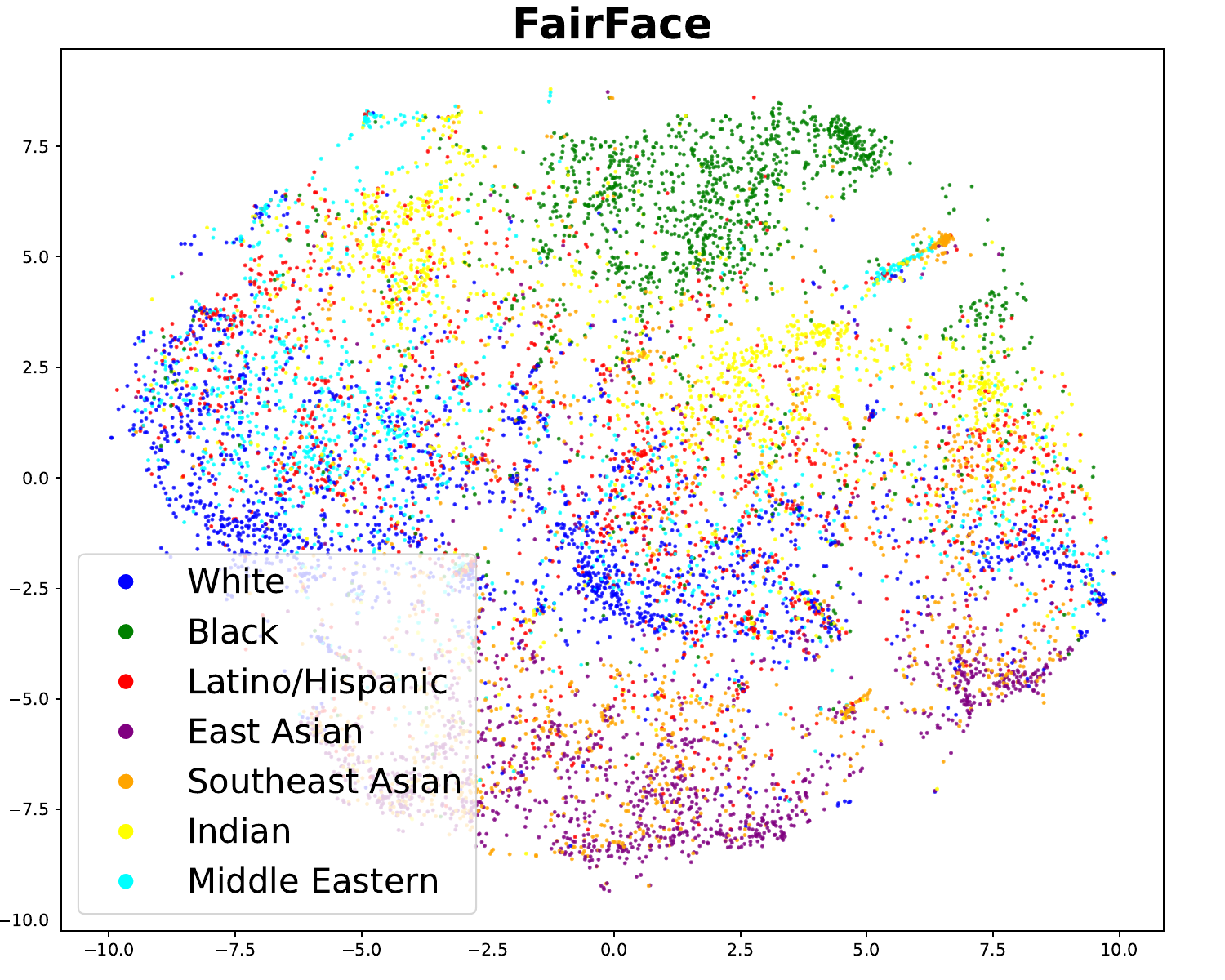}}
        \subfloat{\includegraphics[width=0.248\linewidth]{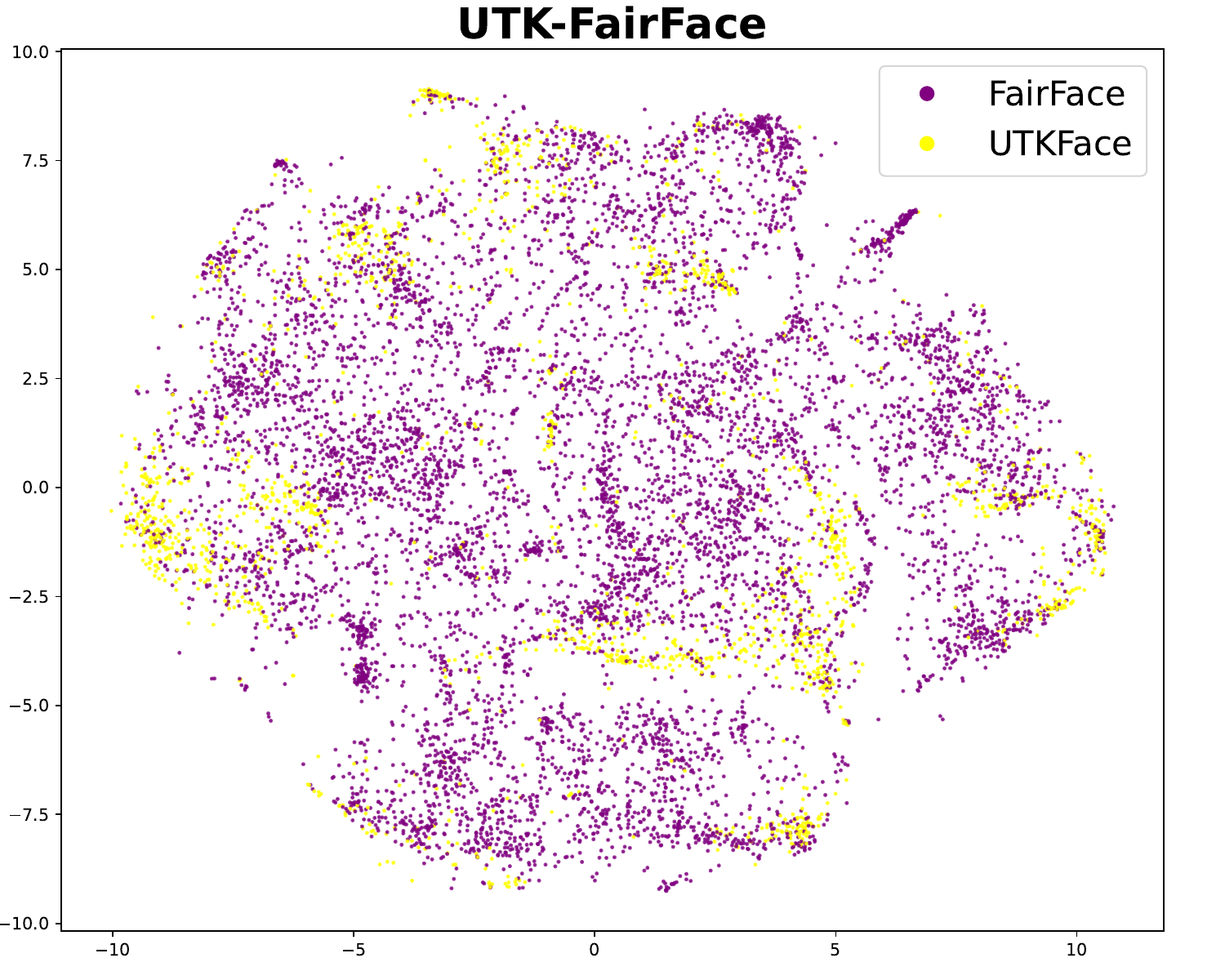}}
        \subfloat{\includegraphics[width=0.210\linewidth]{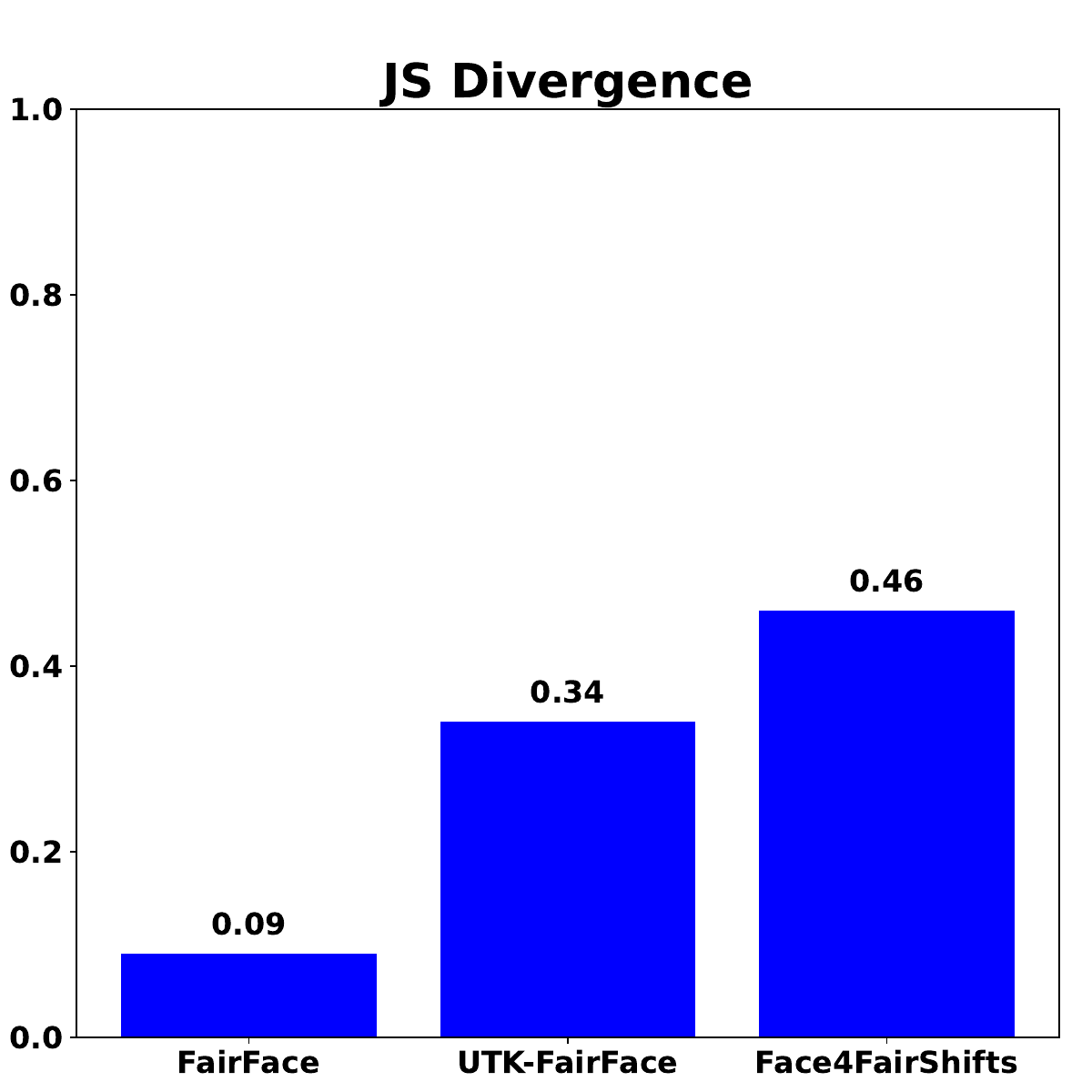}}
    \end{subfigure}
    \vspace{-5mm}
    \caption{t-SNE visualization of datasets for covariate shifts across different domains, pre-trained with \textbf{(Upper)} ResNet50 \cite{he2016deep} and \textbf{(Lower)} ViT-Base \cite{dosovitskiy2020image} on \textsc{ImageNet} \cite{deng2009imagenet}.}
    \label{fig:tsne}
    \vspace{-5mm}
\end{figure}

To assess the effectiveness and robustness of \textsc{\sysname{}}, we conduct extensive experiments across four key machine learning research areas: fairness learning, OOD generalization, OOD detection, and FairOG. 
In each area, we compare \textsc{\sysname{}} with existing datasets that have been adapted for the corresponding problem setting and evaluate it against state-of-the-art methods to ensure a comprehensive benchmark.
More details of experiment settings and results refer to Appendix \ref{sec:app_implementationdetails}.
These experiments aim to address the following questions:
\begin{enumerate}[leftmargin=*]
    \item How does \textsc{\sysname{}} capture and reflect the impact of covariate shifts across different domains compared to existing facial datasets? (\textbf{Answers}: Section \ref{sec:featurespace} and Figure \ref{fig:tsne})
    \item Compared to existing datasets across four different machine learning problems (fairness learning, OOD generalization, OOD detection, and FairOG), does \textsc{\sysname{}} pose greater challenges for state-of-the-art baseline methods, thereby providing more room for improvement and future research? (\textbf{Answers}: Sections \ref{sec:fairlearning}, \ref{sec:oodgen}, \ref{sec:ooddetect}, \ref{sec:fairod} and Tables \ref{tab:fairness_result}, \ref{tab:domain_generalization_result}, \ref{tab:sensory_ood_detection_result}, \ref{tab:intra_domain_detection_result}, \ref{tab:inter_domain_ood_detection_result}, \ref{tab:fairness-aware-domain-generalization_result})
    \item Compared to existing datasets across four different machine learning problems, does \textsc{\sysname{}} lead to more consistent performance across state-of-the-art baseline methods while providing a more stable benchmark for evaluation? (\textbf{Answers}: The $\epsilon$ and $\pm$ indicators in Tables \ref{tab:fairness_result}, \ref{tab:domain_generalization_result}, \ref{tab:sensory_ood_detection_result}, \ref{tab:intra_domain_detection_result}, \ref{tab:inter_domain_ood_detection_result}, \ref{tab:fairness-aware-domain-generalization_result})
\end{enumerate}

\begin{wraptable}{r}{0.655\textwidth}
    \centering
    \tiny
    \setlength\tabcolsep{1pt}
    \vspace{-3mm}
    \caption{Overview of Experimental Datasets}
    \begin{tabular}{ccccccccc}
    \toprule
        \multirow{3}{*}{\textbf{Datasets}} & \multirow{3}{*}{\textbf{Domains}} & \multirow{3}{*}{\textbf{Anno.}} & \multicolumn{6}{c}{\textbf{Tasks}}\\
    \cmidrule{4-9}
         & & & \textbf{FairL} & \textbf{OOD Gen.} & \multicolumn{3}{c}{\textbf{OOD Det.} (Sec. \ref{sec:ooddetect})} & \textbf{FairOG} \\
         & & & (Sec. \ref{sec:fairlearning}) & (Sec. \ref{sec:oodgen}) & \textbf{Sensory} & \textbf{Intra-Sem.} & \textbf{Inter-Sem.} & (Sec. \ref{sec:fairod})\\
    \midrule
        \textsc{CelebA} \cite{CelebA} & - & 40 & \textcolor{green}{\ding{51}} & \textcolor{red}{\ding{55}} & \textcolor{red}{\ding{55}} & \textcolor{green}{\ding{51}} & \textcolor{red}{\ding{55}} & \textcolor{red}{\ding{55}}\\
        \textsc{UTKFace} \cite{utk} & -  & 3 & \textcolor{green}{\ding{51}} & \textcolor{red}{\ding{55}} & \textcolor{red}{\ding{55}} & \textcolor{green}{\ding{51}} & \textcolor{red}{\ding{55}} & \textcolor{red}{\ding{55}} \\
        \textsc{FairFace} \cite{fairface} & 7 (Races) & 3 & \textcolor{green}{\ding{51}} & \textcolor{green}{\ding{51}} & \textcolor{green}{\ding{51}} & \textcolor{green}{\ding{51}} & \textcolor{green}{\ding{51}} & \textcolor{green}{\ding{51}} \\
        \textsc{UTK-FairFace} \cite{utk-fairface} & 2 (Datasets) & 3 & \textcolor{green}{\ding{51}} & \textcolor{green}{\ding{51}} & \textcolor{green}{\ding{51}} & \textcolor{green}{\ding{51}} & \textcolor{green}{\ding{51}} & \textcolor{green}{\ding{51}}\\
    \midrule
        \textsc{\sysname{}} & 4 (P,A,C,S) & 39 & \textcolor{green}{\ding{51}} & \textcolor{green}{\ding{51}} & \textcolor{green}{\ding{51}} & \textcolor{green}{\ding{51}} & \textcolor{green}{\ding{51}} & \textcolor{green}{\ding{51}}\\
    \bottomrule
    \end{tabular}
    \label{tab:expdatasets}
    \vspace{-3mm}
\end{wraptable}

\textbf{Datasets.}
(1) \textsc{CelebA} \cite{CelebA} is a large-scale facial attributes dataset comprising 200,599 celebrity images, each annotated with 40 binary attributes. While it has been widely utilized for fairness learning, it lacks race annotations. Additionally, it has rarely been employed for investigating distribution shifts.
(2) \textsc{UTKFace} \cite{utk} is a facial dataset with a wide age span, consisting of over 20,000 face images annotated with age, gender, and race, including White, Black, Asian, Indian, and Others. Similar to \textsc{CelebA}, this dataset has rarely been utilized for studying distribution shifts. 
(3) \textsc{FairFace} \cite{fairface} is a facial dataset designed to provide a balanced representation across racial groups. It comprises 108,501 images spanning seven racial categories: Black, East Asian, Indian, Latino, Middle Eastern, Southeast Asian, and White. Following the approach in \cite{fedora}, the seven racial groups are treated as distinct domains.
(4) \textsc{UTK-FairFace} \cite{utk-fairface} is a semi-synthetic dataset created by combining \textsc{UTKFace} and \textsc{FairFace}. In \cite{utk-fairface}, the authors argued that a distribution shift exists between the two datasets due to differences in image sources, leading to the consideration of each dataset as a separate domain. However, this dataset primarily relies on source discrepancies rather than naturally occurring domain shifts.
An overview of all datasets used for experiments is given in Table \ref{tab:expdatasets}.

Due to space constraints, we defer the detailed introduction of baseline methods and implementation specifics for each research area (Sections \ref{sec:fairlearning}, \ref{sec:oodgen}, \ref{sec:ooddetect}, and \ref{sec:fairod}) to the Appendix. 
All results presented in Tables \ref{tab:fairness_result}, \ref{tab:domain_generalization_result}, \ref{tab:sensory_ood_detection_result}, \ref{tab:intra_domain_detection_result}, \ref{tab:inter_domain_ood_detection_result}, and \ref{tab:fairness-aware-domain-generalization_result} are averaged over three repeated runs.


\subsection{Feature Space Analysis}
\label{sec:featurespace}
We perform a preliminary analysis to contrast the covariate shifts within our \textsc{\sysname{}} to that of prior used facial datasets for FairOG, such as \textsc{FairFace} in \cite{fedora} and \textsc{UTK-Fairface} in \cite{utk-fairface}.
Figure \ref{fig:tsne} presents the t-SNE visualization of data samples across different domains, utilizing ResNet50 \cite{he2016deep} and ViT-Base \cite{dosovitskiy2020image} as feature extraction backbones pre-trained on \textsc{ImageNet} \cite{deng2009imagenet}.
Additionally, we quantify the covariate shift across domains for all datasets using the Jensen-Shannon (JS) divergence \cite{endres2003new}, which is computed based on the Kullback-Leibler (KL) divergence.
Our results clearly demonstrate that the covariate shift in \textsc{\sysname{}} is significantly higher compared to other datasets, highlighting the dataset's effectiveness in modeling real-world distributional variations.

\subsection{Fairness Learning}
\label{sec:fairlearning}
\textbf{Setting.}
To evaluate model fairness, we treat each dataset as a whole, disregarding domain distinctions. Specifically, we consider two label-sensitive settings: age-gender and age-race. In these settings, age is discretized into elderly and non-elderly, gender is categorized as male and female, and race is grouped into Black and non-Black. This approach allows us to assess fairness disparities across sensitive attributes without the influence of domain shifts.

\textbf{Evaluation Metrics.}
We employ two widely used fairness evaluation metrics: the absolute difference of demographic difference ($\Delta$DP) \cite{dwork2012fairness} and that of equalized odds ($\Delta$EO) \cite{hardt2016equality}.
A value of zero for these metrics indicates fair predictions, meaning the model's decisions are independent of the sensitive attribute $Z$.
Additionally, to analyze the stability of these baselines on a given dataset, we introduce two stability indicators, $\epsilon$ and $\pm$. $\epsilon$ quantifies the average absolute difference between any two baseline methods.
The $\pm$ indicator represents the standard deviation of the performance across all baseline methods.  
Smaller values of both $\epsilon$ and $\pm$ indicate more stable and consistent performance across different baselines. 

\textbf{Baseline Methods.} 
We evaluate the performance of all datasets using five widely adopted baseline methods, including LFR~\cite{lfr}, GSR~\cite{gsr}, AD~\cite{ad}, CSAD~\cite{csad}, and FNF~\cite{fnf}.


\textbf{Results.}
Table \ref{tab:fairness_result} illustrates that \textsc{\sysname{}} presents greater challenges compared to all other datasets, as evidenced by the lowest accuracy and the highest values of $\Delta$DP and $\Delta$EO across the majority of baseline methods. 
Table \ref{tab:appendix_fairness_result} in the Appendix provides a more detailed version of Table \ref{tab:fairness_result}, including standard deviations for each baseline method.
Furthermore, \textsc{\sysname{}} exhibits more stable performance across baseline methods, as indicated by the consistently lower values of $\epsilon$ and $\pm$ across all metrics. Note that since \textsc{CelebA} does not include race attributes, it is excluded from the age-race setting in our experiments.

\begin{table*}[ht]
\centering
\tiny
\setlength\tabcolsep{0.75pt}
\caption{Performance of Fairness Learning Using Multiple Baselines.}
\vspace{-3mm}
\begin{threeparttable}
\begin{tabular}{c|c|c|c|c|c|c|c|c|c|c|c|c|c|c|c|c|c|c|c|c|c|c}
\toprule
 \textbf{Label} & \multirow{2.4}{*}{\textbf{Dataset}}& \multicolumn{7}{c|}{\textbf{Accuracy}}& \multicolumn{7}{c|}{\textbf{$\Delta\text{DP}$}} & \multicolumn{7}{c}{\textbf{$\Delta\text{EO}$}}\\

    \cmidrule(lr){1-1}
  \cmidrule(lr){3-9} \cmidrule(lr){10-16} \cmidrule(lr){17-23}
      \textbf{Sen.}&& LFR &  GSR & AD &
      CSAD &
      FNF & $\epsilon$ & $\pm$ & LFR &  GSR & AD &
      CSAD &
      FNF & $\epsilon$ & $\pm$ & LFR &  GSR & AD &
      CSAD &
      FNF & $\epsilon$ & $\pm$\\

    \midrule

 \multirow{5}{*}{\rotatebox{90}{\makecell[c]{\textbf{Age}\\\textbf{Gender}}}}&\textsc{Celeba}&78.92 & 80.49 & 78.48 & 85.37 & 85.96 & 0.043 & 0.036 & 0.062 & 0.069 & 0.067 & 0.049 & 0.047 & 0.012 & 0.010 & 0.062 & 0.051 & 0.051 & 0.036 & 0.036 & 0.013 & 0.011\\

&\textsc{UTKFace}&78.13 & 78.27 & 78.93 & 83.29 & 83.50 & 0.032 & 0.027 & 0.021 & 0.020 & 0.021 & 0.020 & 0.019 & \textbf{0.001} & \textbf{0.001} & 0.063 & 0.070 & 0.070 & 0.053 & 0.052 & 0.011 & 0.009\\
 &\textsc{FairFace}&87.32 & 83.21 & 83.87 & 89.25 & 89.69 & 0.037 & 0.030 & 0.071 & 0.025 & 0.045 & 0.025 & 0.026 & 0.023 & 0.020 & 0.083 & 0.064 & 0.065 & 0.059 & 0.060 & 0.010 & 0.010\\

 &\textsc{\utkfair{}}&86.43 & 79.38 & 78.40 & 84.44 & 83.87 & 0.042 & 0.034 & 0.045 & 0.019 & 0.021 & 0.018 & 0.018 & 0.011 & 0.011 & 0.047 & 0.040 & 0.042 & 0.033 & 0.030 & 0.009 & \textbf{0.007}\\
 &\textsc{\sysname{}}&\textbf{75.06} & \textbf{74.50} & \textbf{75.90} & \textbf{78.49} & \textbf{79.28} & \textbf{0.026} & \textbf{0.021} & \textbf{0.092} & \textbf{0.083} & \textbf{0.083} & \textbf{0.079} & \textbf{0.071} & 0.009 & 0.008 & \textbf{0.086} & \textbf{0.083} & \textbf{0.083} & \textbf{0.074} & \textbf{0.070} & \textbf{0.008} & \textbf{0.007}\\

 \midrule

 \multirow{4}{*}{\rotatebox{90}{\makecell[c]{\textbf{Age}\\\textbf{Race}}}}&\textsc{UTKFace}&76.90 & 76.37 & 75.44 & 78.81 & 77.24 & 0.015 & 0.012 & 0.096 & 0.121 & 0.100 & 0.083 & 0.097 & 0.016 & 0.014 & 0.098 & 0.088 & 0.089 & 0.084 & 0.087 & \textbf{0.006} & \textbf{0.005}\\

 &\textsc{FairFace}&87.42 & 84.23 & 82.71 & 88.90 & 88.89 & 0.034 & 0.028 & 0.144 & 0.134 & 0.129 & 0.126 & 0.123 & 0.010 & 0.008 & 0.129 & 0.120 & 0.123 & 0.113 & 0.119 & 0.007 & 0.006\\

 &\textsc{\utkfair{}}&84.67 & 81.63 & 81.62 & 85.60 & 85.19 & 0.023 & 0.020 & 0.122 & 0.094 & 0.098 & 0.092 & 0.095 & 0.013 & 0.013 & 0.102 & 0.086 & 0.088 & 0.086 & 0.087 & 0.007 & 0.007\\
 &\textsc{\sysname{}}&\textbf{73.53} & \textbf{73.97} & \textbf{74.70} & \textbf{76.37} & \textbf{75.30} & \textbf{0.014} & \textbf{0.011} & \textbf{0.154} & \textbf{0.150} & \textbf{0.155} & \textbf{0.145} & \textbf{0.143} & \textbf{0.006} & \textbf{0.005} & \textbf{0.142} & \textbf{0.136} & \textbf{0.142} & \textbf{0.136} & \textbf{0.132} & \textbf{0.006} & \textbf{0.005}\\

\bottomrule
\end{tabular}
\end{threeparttable}

\label{tab:fairness_result}
\vspace{-3mm}
\end{table*}

\begin{table*}[ht]
\centering
\tiny
\setlength\tabcolsep{2pt}
\caption{Average Performance of OOD Generalization across Domains.}
\vspace{-3mm}
\begin{threeparttable}
\begin{tabular}{c|c|c|c|c|c|c|c|c|c|c|c|c|c|c|c|c|c}
\toprule
 \multirow{2.4}{*}{\textbf{Label}} & \multirow{2.4}{*}{\textbf{Dataset}}& \multicolumn{8}{c|}{\textbf{Accuracy}}& \multicolumn{8}{c}{\textbf{F-1 Score}}\\

  \cmidrule(lr){3-10} \cmidrule(lr){11-18}
 
      &&ERM &  IRM & GDRO &
      Mixup & MMD &
      MBDG & $\epsilon$ & $\pm$ &ERM &  IRM & GDRO &
      Mixup & MMD &
      MBDG & $\epsilon$ & $\pm$ \\

    \midrule

 \multirow{3}{*}{\rotatebox{0}{\makecell[c]{\textbf{Age}}}}&\textsc{FairFace}&92.35 & 92.19 & 88.68 & 86.38 & 92.46 & 94.46 & 0.035 & 0.030 & 63.92 & 63.03 & 16.31 & \textbf{2.60} & 50.66 & 65.93 & 0.315 & 0.274\\

 &\textsc{\utkfair{}}&90.20 & 89.32 & 88.47 & \textbf{81.31} & 89.43 & 91.34 & 0.037 & 0.036 & 62.07 & 61.83 & 9.32 & 14.14 & 58.25 & 63.93 & 0.280 & 0.258\\
 &\textsc{\sysname{}} &\textbf{87.99} & \textbf{88.36} & \textbf{88.21} & 88.41 & \textbf{88.22} & \textbf{90.23} & \textbf{0.008} & \textbf{0.008} & \textbf{25.71} & \textbf{35.49} & \textbf{3.75} & 5.95 & \textbf{46.32} & \textbf{41.29} & \textbf{0.219} & \textbf{0.181}\\

 \midrule

 \multirow{3}{*}{\rotatebox{0}{\makecell[c]{\textbf{Gender}}}}&\textsc{FairFace}&87.90 & 87.71 & 57.11 & 49.85 & 52.78 & 90.85 & 0.227 & 0.196 & 86.29 & 86.36 & 46.38 & 47.27 &\textbf{ 14.14} & 87.80 & 0.351 & 0.303\\

 &\textsc{\utkfair{}}&85.00 & 83.77 & 57.37 & \textbf{47.30} & 47.92 & 88.33 & 0.229 & 0.195 & 85.10 & 81.92 & 44.66 & \textbf{41.12} & 24.66 & 83.92 & 0.312 & 0.265\\
 &\textsc{\sysname{}}&\textbf{79.00} &\textbf{ 77.58} & \textbf{55.75} & 51.09 & \textbf{46.31} & \textbf{85.26} & \textbf{0.200} &\textbf{ 0.167} &\textbf{ 81.62} & \textbf{79.85 }& \textbf{43.45} & 41.99 & 32.25 & \textbf{82.45} &\textbf{ 0.271 }& \textbf{0.234}\\




\bottomrule
\end{tabular}
\end{threeparttable}

\label{tab:domain_generalization_result}
\vspace{-5mm}
\end{table*}

\subsection{OOD Generalization}
\label{sec:oodgen}
\textbf{Setting.}
In OOD generalization, since target domains are unknown and inaccessible during training, followed \cite{mbdg,gulrajani2020search}, we use leave-one-domain-out validation criteria. Specifically, we evaluate selected datasets, \textsc{FairFace}, \textsc{UTK-Fairface}, and \textsc{\sysname{}}, by holding out one domain during training and assessing model performance on the held-out one (target). The final performance is reported as the average across all held-out domains, ensuring a robust evaluation of generalization under distribution shifts.

\textbf{Baseline Methods.}
The performance of all datasets is evaluated using six widely recognized methods ERM~\cite{erm}, IRM~\cite{irm}, GDRO~\cite{gdro}, Mixup~\cite{mixup}, MMD~\cite{mmd}, MBDG~\cite{mbdg}. These methods are implemented in the DomainBed repository \cite{gulrajani2020search}, a widely used benchmarking framework for domain generalization, which has gained significant traction, accumulating over 1.5k stars on GitHub.

\textbf{Results.}
Similar to the fairness learning results in Section \ref{sec:fairlearning}, Table \ref{tab:domain_generalization_result} highlights that \textsc{\sysname{}} presents greater challenges compared to other datasets, as indicated by the lowest accuracy and F-1 scores across various baseline methods. Additionally, it demonstrates more stable performance, evidenced by the consistently smaller values of $\epsilon$ and $\pm$.



\begin{table*}[t]
\centering
\tiny
\setlength\tabcolsep{0.5pt}
\caption{Average Performance of Inter-Domain Sensory OOD Detection across Domains using Age as Class Labels.}
\vspace{-3mm}
\begin{threeparttable}
\begin{tabular}{l|c|c|c|c|c|c|c|c|c|c|c|c|c|c|c|c|c|c|c|c|c}
\toprule
  \multirow{2.4}{*}{\textbf{Dataset}}& \multicolumn{7}{c|}{\textbf{AUROC}}& \multicolumn{7}{c|}{\textbf{AUPR}} & \multicolumn{7}{c}{\textbf{InD/OOD Accuracy} (Reference Only)}\\

  \cmidrule(lr){2-8} \cmidrule(lr){9-15} \cmidrule(lr){16-22}
      &ocSVM &  DDU & MSP &
      Energy &
      Entropy & $\epsilon$ & $\pm$ &ocSVM &  DDU & MSP &
      Energy &
      Entropy & $\epsilon$ & $\pm$ &ocSVM &  DDU & MSP &
      Energy &
      Entropy & $\epsilon$ & $\pm$\\

    \midrule

 FairFace&50.13 & 52.39 & 51.27 & 50.83 & 50.04 & 0.012 & 0.010 & 12.83 & 22.91 & 17.63 & 16.36 & 13.83 & 0.053 & 0.042 & 50.13 & 51.77 & 52.03 & 50.85 & 50.01 & 0.011 & 0.009\\

 \utkfair{}&52.29 & 54.91 & 53.83 & 54.57 & 51.85 & 0.017 & 0.014 & 18.27 & 24.83 & 22.77 & 26.66 & 15.27 & 0.059 & 0.047 & 51.63 & 53.59 & 52.00 & 53.34 & 51.36 & 0.012 & 0.010\\
 \sysname{}&64.29 & 66.28 & 66.01 & 65.53 & 64.77 & 0.010 & 0.008 & 45.26 & 51.25 & 48.13 & 50.03 & 47.28 & 0.029 & 0.023 & 60.02 & 61.68 & 61.17 & 60.90 & 60.31 & 0.008 & 0.007\\

\bottomrule
\end{tabular}
\end{threeparttable}

\label{tab:sensory_ood_detection_result}
\vspace{-3mm}
\end{table*}



\begin{table*}[t]
\centering
\tiny
\setlength\tabcolsep{0.5pt}
\caption{Performance of Intra-Domain Semantic OOD Detection using Age as Class Labels.}
\vspace{-3mm}
\begin{threeparttable}
\begin{tabular}{l|c|c|c|c|c|c|c|c|c|c|c|c|c|c|c|c|c|c|c|c|c}
\toprule
  \multirow{2.4}{*}{\textbf{Dataset}}& \multicolumn{7}{c|}{\textbf{AUROC}}& \multicolumn{7}{c|}{\textbf{AUPR}} & \multicolumn{7}{c}{\textbf{InD Classification Accuracy} (Reference Only)}\\

  \cmidrule(lr){2-8} \cmidrule(lr){9-15} \cmidrule(lr){16-22}
      &ocSVM &  DDU & MSP &
      Energy &
      Entropy & $\epsilon$ & $\pm$ &ocSVM &  DDU & MSP &
      Energy &
      Entropy & $\epsilon$ & $\pm$ &ocSVM &  DDU & MSP &
      Energy &
      Entropy & $\epsilon$ & $\pm$\\

    \midrule

 FairFace&78.33 & 95.89 & 92.40 & 92.06 & 84.08 & 0.087 & 0.072 & \textbf{60.12} & 82.15 & 77.04 & 76.45 & 73.08 & 0.096 & 0.083 & 69.40 & 79.67 & 77.48 & 74.01 & 76.23 & 0.048 & 0.039\\

 \utkfair{}&88.06 & 84.68 & 89.63 & 93.92 & 87.02 & 0.042 & 0.034 & 64.07 & 60.28 & 77.68 & 84.80 & 70.32 & 0.125 & 0.100 & 75.00 & 84.23 & 69.07 & \textbf{61.03} & 67.10 & 0.109 & 0.088\\
 
 \sysname{}&\textbf{71.11} & \textbf{74.76} & \textbf{72.73} & \textbf{69.81} & \textbf{70.25} & \textbf{0.025} & \textbf{0.020} & 61.09 & \textbf{57.72} & \textbf{56.51} & \textbf{66.37} & \textbf{54.90} & \textbf{0.055} & \textbf{0.046} & \textbf{55.45} & \textbf{61.27} & \textbf{57.16} & 61.59 & \textbf{57.42} & \textbf{0.033} & \textbf{0.027}\\

\bottomrule
\end{tabular}
\end{threeparttable}

\label{tab:intra_domain_detection_result}
\vspace{-3mm}
\end{table*}




\begin{table*}[t!]
\centering
\tiny
\setlength\tabcolsep{0pt}
\caption{Average Performance of Inter-Domain Semantic OOD Detection across Domains using Age as Class Labels.}
\vspace{-3mm}
\begin{threeparttable}
\begin{tabular}{l|c|c|c|c|c|c|c|c|c|c|c|c|c|c|c|c|c|c|c|c|c}
\toprule
  \multirow{2.4}{*}{\textbf{Dataset}}& \multicolumn{7}{c|}{\textbf{AUROC}}& \multicolumn{7}{c|}{\textbf{AUPR}} & \multicolumn{7}{c}{\textbf{InD Classification Accuracy} (Reference Only)}\\

  \cmidrule(lr){2-8} \cmidrule(lr){9-15} \cmidrule(lr){16-22}
      &EDst &  SCONE & DAML &
      MEDIC &
      MADOD& $\epsilon$ & $\pm$ &EDst &  SCONE & DAML &
      MEDIC &
      MADOD& $\epsilon$ & $\pm$ &EDst &  SCONE & DAML &
      MEDIC &
      MADOD& $\epsilon$ & $\pm$\\

    \midrule

 FairFace&70.28 & 71.28 & 69.28 & 71.93 & 74.25 & 0.023 & 0.019 & 32.48 & 36.27 & 41.28 & 44.27 & 47.28 & 0.075 & 0.060 & 66.00 & 62.94 & 61.76 & 59.28 & 61.27 & 0.030 & 0.025\\

 \utkfair{}&75.22 & 72.22 & 77.28 & 80.29 & 80.14 & 0.042 & 0.034 & 48.27 & 44.42 & 49.28 & 39.20 & 50.13 & 0.053 & 0.045 & 71.29 & 72.15 & 77.28 & 75.27 & 75.81 & 0.031 & 0.025\\
 \sysname{}&\textbf{55.16} & \textbf{56.13} & \textbf{55.28} & \textbf{57.28} & \textbf{57.18} & \textbf{0.012} & \textbf{0.010} & \textbf{27.18} & \textbf{26.94} & \textbf{25.81} & \textbf{28.92} & \textbf{29.38} & \textbf{0.018} & \textbf{0.015} & \textbf{52.38} & \textbf{52.39} & \textbf{52.59} & \textbf{54.02} & \textbf{53.02} & \textbf{0.008} & \textbf{0.007}\\

\bottomrule
\end{tabular}
\end{threeparttable}

\label{tab:inter_domain_ood_detection_result}
\vspace{-3mm}
\end{table*}

\subsection{OOD Detection}
\label{sec:ooddetect}
\textbf{Setting.}
In OOD detection, we consider three distinct scenarios, as illustrated in Figure \ref{fig:ood_detection} in the Appendix.
In inter-domain sensory OOD detection, images exhibiting covariate shifts are considered OOD.
In intra-domain and inter-domain semantic OOD detection, images with semantic shifts are identified as OOD. 
Specifically, in inter-domain semantic OOD detection, only images with covariate shifts are regarded as in-distribution (InD), while those with semantic shifts are classified as OOD.
Detailed experiment settings are provided in Section \ref{sec:app_ooddetection} of the Appendix.

\textbf{Evaluation Metrics.}
We evaluate OOD detection performance across all scenarios using AUROC \cite{davis2006relationship} and AUPR \cite{manning1999foundations}.
AUROC measures the trade-off between the true positive rate and the false positive rate, providing an overall assessment of a model’s ability to distinguish between InD and OOD samples. 
AUPR evaluates performance based on precision and recall, which are particularly useful in imbalanced settings where OOD samples are rare. 
Notice that the accuracy (InD/OOD for sensory, InD for intra- and inter-domain semantic) is provided for reference only.

\textbf{Baseline Methods.}
For inter-domain sensory OOD detection and intra-domain semantic OOD detection, we use One-class SVM (ocSVM)~\cite{oc-svm}, DDU~\cite{ddu}, MSP~\cite{msp}, Energy~\cite{energy} and Entropy~\cite{entropy} as baseline methods. 
For inter-domain semantic OOD detection, we use EDst~\cite{edst}, SCONE~\cite{scone}, DAML~\cite{daml}, MEDIC~\cite{medic}, MADOD~\cite{maood} as baseline methods.
These methods provide a diverse range of approaches for evaluating OOD detection performance under both covariate and semantic distribution shifts.

\textbf{Results}
Table \ref{tab:sensory_ood_detection_result} shows that \textsc{\sysname{}} achieves the best performance in inter-domain sensory OOD detection, outperforming \textsc{FairFace} and \textsc{UTK-FairFace} in terms of AUROC, AUPR, and InD/OOD accuracy. In this setting, images exhibiting covariate shifts are treated as OOD. The results in this table align with the trends observed in Figure \ref{fig:tsne}, further supporting the effectiveness of \textsc{\sysname{}}.
Additionally, Tables \ref{tab:intra_domain_detection_result} and \ref{tab:inter_domain_ood_detection_result} present the results for the settings of intra- and inter-domain OOD detection. Again, \textsc{\sysname{}} poses challenges across all settings compared to \textsc{FairFace} and \textsc{UTK-FairFace}, demonstrating consistently lower performance across different baseline methods while maintaining stability across them.

\vspace{-3mm}
\subsection{Fairness-aware OOD Generalization}
\label{sec:fairod}
\textbf{Setting.}
Similar to the settings used in fairness learning and OOD generalization, we evaluate \textsc{FairFace}, \textsc{UTK-FairFace}, and our \textsc{\sysname{}} across multiple domains in the FairOG setting. We consider two label-sensitive settings to examine fairness disparities while accounting for the impact of domain shifts. This approach provides a comprehensive evaluation of how fairness-aware models generalize under distributional variations. To ensure a robust assessment, the final performance is reported as the average across all held-out domains.

\textbf{Baseline Methods.}
We evaluate all datasets using six recently published methods, FCR~\cite{utk-fairface}, EIIL~\cite{eiil}, FVAE~\cite{fvae}, FATDM~\cite{fatdm}, FLAIR~\cite{flair}, and FEDORA~\cite{fedora}.

\textbf{Results.}
Since the race attribute is used to define domains in \cite{fedora}, we exclude \textsc{FairFace} from the age-race setting in this experiment.
Table \ref{tab:fairness-aware-domain-generalization_result} further demonstrates that \textsc{\sysname{}} presents greater challenges compared to other datasets under both settings while maintaining stable performance across various state-of-the-art methods.

\begin{table*}[t!]
\centering
\tiny
\setlength\tabcolsep{0pt}
\caption{Average Performance of Fairness-aware OOD Generalization across Domains.}
\vspace{-3mm}
\begin{threeparttable}
\begin{tabular}{c|c|c|c|c|c|c|c|c|c|c|c|c|c|c|c|c|c|c|c|c|c|c|c|c|c}
\toprule
 \textbf{Label} & \multirow{2.6}{*}{\textbf{Dataset}}& \multicolumn{8}{c|}{\textbf{Accuracy}}& \multicolumn{8}{c|}{\textbf{$\Delta\text{DP}$}} & \multicolumn{8}{c}{\textbf{$\Delta\text{EO}$}}\\

  \cmidrule(lr){1-1} \cmidrule(lr){3-10} \cmidrule(lr){11-18} \cmidrule(lr){19-26}

      \textbf{Sen.} &&FCR &  EIIL & FVAE &
      FATDM &
      FLAIR&FEDORA& $\epsilon$ & $\pm$ &FCR &  EIIL & FVAE &
      FATDM &
      FLAIR&FEDORA& $\epsilon$ & $\pm$ &FCR &  EIIL & FVAE &
      FATDM &
      FLAIR&FEDORA& $\epsilon$ & $\pm$\\

    \midrule

 \multirow{3}{*}{\rotatebox{90}{\makecell[c]{\textbf{Age}\\\textbf{Gender}}}}&\textsc{FairFace}&87.41 & 82.12 & 86.31 & 91.32 & 91.73 & 92.42 & 0.048 & 0.040 & 0.102 & \textbf{0.151} & 0.132 & 0.112 & 0.101 & 0.105 & 0.023 & 0.020 & 0.133 & 0.140 & 0.136 & 0.122 & 0.112 & 0.102 & 0.018 & 0.015\\

 &\textsc{\utkfair{}}&84.21 & 80.27 & 83.24 & 89.12 & 88.27 & 90.12 & 0.047 & 0.039 & 0.082 & 0.120 & 0.124 & 0.083 & 0.092 & 0.090 & 0.021 & 0.019 & 0.119 & 0.120 & 0.143 & 0.125 & 0.129 & 0.102 & 0.016 & 0.013\\
 &\textsc{\sysname{}}&\textbf{78.91} & \textbf{76.13} & \textbf{74.62} & \textbf{81.44} & \textbf{82.25} & \textbf{82.22} & \textbf{0.039} & \textbf{0.033} & \textbf{0.142} & 0.149 & \textbf{0.151} & \textbf{0.133} & \textbf{0.130} & \textbf{0.125} & \textbf{0.013} & \textbf{0.011} & \textbf{0.143} & \textbf{0.152} & \textbf{0.154} & \textbf{0.152} & \textbf{0.148} & \textbf{0.132} & \textbf{0.009} & \textbf{0.008}\\

 \midrule

 \multirow{2}{*}{\rotatebox{90}{\makecell[c]{\textbf{Age}\\\textbf{Race}}}}&\textsc{\utkfair{}}&79.08 & 72.83 & 78.91 & 78.29 & 77.22 & 78.52 & 0.024 & 0.024 & 0.143 & 0.103 & 0.173 & 0.154 & 0.143 & 0.140 & 0.026 & 0.023 & 0.142 & 0.113 & 0.173 & 0.133 & 0.143 & 0.133 & 0.023 & 0.020\\

 &\textsc{\sysname{}}&\textbf{72.25} & \textbf{69.13} & \textbf{72.82} & \textbf{73.28} & \textbf{74.22} & \textbf{74.53} & \textbf{0.022} & \textbf{0.019} & \textbf{0.173} & \textbf{0.169} & \textbf{0.179} & \textbf{0.163} & \textbf{0.161} & \textbf{0.160 }& \textbf{0.009} & \textbf{0.008} & \textbf{0.169} & \textbf{0.153} & \textbf{0.177} & \textbf{0.145} & \textbf{0.151} & \textbf{0.146} & \textbf{0.015} & \textbf{0.013}\\

\bottomrule
\end{tabular}
\end{threeparttable}

\label{tab:fairness-aware-domain-generalization_result}
\vspace{-3mm}
\end{table*}

\section{Conclusion}
\label{sec:conclusion}
\vspace{-3mm}
In this paper, we introduce \textsc{\sysname{}}, a large-scale facial image benchmark designed to study fairness and robustness under domain shifts. It spans four visual domains, comprising 100,000 images with 42 annotations across 15 attributes. Extensive experiments demonstrate that \textsc{\sysname{}} serves as a valuable resource for developing and evaluating fair and robust machine learning techniques. 

\clearpage
\bibliographystyle{abbrv}
\bibliography{reference}

\clearpage
\appendix


\section{Related Work}
\label{sec:relatedwork}
Table~\ref{tab:facedatasets} presents a list of related real-world image datasets focused on facial attributes and distribution shifts.

\textbf{Datasets for Fairness Learning.}
Fairness-aware machine learning has been extensively studied using both tabular and image datasets that capture demographic disparities.
Several widely used tabular datasets, such as \textsc{Adult Income} \cite{adultincome}, \textsc{COMPAS} \cite{compas}, \textsc{German Credit} \cite{germancredit}, and \textsc{Bank Marketing} \cite{bankmarketing}, are frequently used to evaluate fairness learning algorithms due to their inclusion of sensitive attributes like race and gender. 
In facial image datasets, studies often rely on \textsc{CelebA}~\cite{CelebA}, which provides extensive facial attribute annotations but is imbalanced in demographic representation. 
More balanced alternatives, including \textsc{UTKFace}~\cite{utk} and \textsc{FairFace}~\cite{fairface}, offer a diverse distribution of race, age, and gender, making them more suitable for fairness-aware classification. 
Another dataset, \textsc{WaterBird}~\cite{sagawa2019distributionally}, has been used to study spurious correlations between background features and class labels, serving as an example of how dataset biases can affect classification models.
However, while these datasets facilitate fairness research, they lack inherent covariate shifts and thus do not adequately support studies on FairOG. 

\textbf{Datasets for Distribution Shifts.}
To study distribution shifts, researchers have leveraged datasets designed to capture variations in visual domains. 
Multi-domain benchmarks such as \textsc{VLCS} \cite{torralba2011unbiased}, which combines images from existing datasets \cite{everingham2010pascal,torralba2010labelme,fei2004learning,choi2010exploiting}, introduce natural distribution shifts for object recognition. 
Similarly, \textsc{PACS} \cite{li2017deeper}, containing images categorized into Photo, Art, Cartoon, and Sketch domains, is widely used to evaluate robustness to style variations. 
Additionally, datasets like \textsc{Office} \cite{saenko2010adapting}, composed of images collected from sources such as Amazon, DSLR, and Webcam, enable research on domain adaptation in classification tasks. 
For large-scale multi-domain learning, \textsc{DomainNet} \cite{peng2019moment} provides images from six diverse domains, making it a comprehensive resource for studying robustness across visual shifts. 
Beyond visual style and object recognition, datasets included in \textsc{WILDS} \cite{koh2021wilds}, such as \textsc{Camelyon17} \cite{koh2021wilds} for histopathology analysis and \textsc{FMoW} \cite{koh2021wilds} for satellite imagery, have been used to evaluate distribution shifts in medical and remote sensing applications. 
Although these datasets effectively model covariate and semantic shifts, they do not explicitly include fairness-sensitive attributes, limiting their application in FairOG. 

\textbf{Datasets for Fairness-aware Distribution Shifts.}
FairOG remains an emerging research area, yet existing datasets rarely address both fairness and robustness under distribution shifts.
Some studies~\cite{utk-fairface,fedora,flair} have attempted to modify fairness-focused datasets such as \textsc{UTKFace} and \textsc{FairFace} by partitioning them into synthetic domains or manipulating sample distributions to simulate artificial shifts. 
However, these modifications, such as using the combined \textsc{UTK-FairFace}~\cite{utk-fairface} dataset or partitioning \textsc{FairFace} into seven domains based on race \cite{fedora}, often fail to capture meaningful domain-specific covariate shifts or establish strong correlations between sensitive attributes and class labels.
The absence of a dedicated dataset that simultaneously encapsulates fairness-sensitive attributes and realistic distribution shifts underscores the need for a benchmark tailored to FairOG. \textsc{\sysname{}} addresses these limitations by naturally incorporating both fairness-sensitive attributes and diverse domain shifts, providing a well-structured benchmark for studying FairOG.

\begin{table*}[t]
\tiny
\setlength\tabcolsep{1pt}
    \centering
    \caption{Overview of Related Real-World Image Datasets.}
    \begin{tabular}{lccccccc}
    \toprule
         \multirow{2}{*}{\textbf{Datasets}} & \multirow{2}{*}{\textbf{Year}} & \multirow{2}{*}{\textbf{Data Sources}} & \multirow{2}{*}{\textbf{Dataset Size}} & \multirow{2}{*}{\textbf{\# Annotations}} & \textbf{Sensitive} & \multirow{2}{*}{\textbf{\# Domains}} & \multirow{2}{*}{\textbf{Domains}} \\
         & & & & & \textbf{Attribute} & & \\
    \midrule
      \multicolumn{8}{c}{\textbf{Face Attribute}} \\
      \midrule
        \textsc{MORPH} \cite{ricanek2006morph} & 2006 & Public Data & 55K & 7&Yes & - & - \\
         \midrule
         \textsc{PubFig} \cite{kumar2011describable} & 2011 & Celebrity & 13K &73&Yes& - & - \\
         \midrule
         \textsc{CACD} \cite{chen2015face} & 2015 & Celebrity & 160K & 9&Yes & - & - \\
         \midrule
         \textsc{LFWA+} \cite{CelebA} & 2015 & \textsc{LFW} \cite{huang2008labeled} & 13K &40 &Yes & - & - \\
         \midrule
         \textsc{CelebA} \cite{CelebA} & 2015 & \textsc{CelebFace} \cite{sun2013hybrid,sun2014deep}, \textsc{LFW} \cite{huang2008labeled} & 202K  & 40 & Yes & - & - \\
         \midrule
         \textsc{IMDB-WIKI} \cite{rothe2018deep} & 2016 & IMDB, WIKI & 500K  & 8&Yes & - & - \\
         \midrule
         \textsc{FotW} \cite{escalera2016chalearn} & 2016 & Flickr & 25K  &10 & Yes& - & -\\
         \midrule
         \textsc{UTKFace} \cite{utk} & 2017 & MORPH \cite{ricanek2006morph}, CACD \cite{chen2015face}, Web & 20K &3 & Yes& - & - \\
         \midrule
         \textsc{PPB} \cite{buolamwini2018gender} & 2018 & Government Official Profiles & 1K &2 &Yes & - & - \\
         \midrule
         \textsc{LFW+} \cite{han2017heterogeneous} & 2018 & \textsc{LFW} \cite{huang2008labeled} & 15K &3 &Yes & - & - \\
         \midrule
         \textsc{DiF} \cite{merler2019diversity} & 2019 & Flickr & 1M &10 & Yes& - & -\\
         \midrule
         \textsc{CelebAMask-HQ} \cite{CelebAMask-HQ} & 2020 &\textsc{CelebA-HQ}~\cite{HQ} & 30K &19&Yes & - & - \\
         \midrule
         \textsc{CelebA-Spoof} \cite{CelebA-Spoof} & 2020 &\textsc{CelebA} \cite{CelebA} & 625K & 43 &Yes & - & - \\
         \midrule
         \textsc{CelebA-Dialog} \cite{jiang2021talk,jiang2023talk} & 2021 & \textsc{CelebA} \cite{CelebA} & 30K & 45 &Yes  & - & - \\
    \midrule
      \multicolumn{8}{c}{\textbf{Distribution Shifts}} \\
      \midrule
        \textsc{Office} \cite{saenko2010adapting} & 2010 & Web, Digital SLR Camera, Webcam &4K & No& No& 3 & Amazon, Webcam, DSLR \\
         \midrule
         
         \multirow{2}{*}{\textsc{VLCS} \cite{torralba2011unbiased}} & \multirow{2}{*}{2011} & \textsc{VOC2007} \cite{everingham2010pascal}, \textsc{LabelMe} \cite{torralba2010labelme} & \multirow{2}{*}{10K} & \multirow{2}{*}{No} & \multirow{2}{*}{No} & \multirow{2}{*}{4} & Caltech101, LabelMe\\
        
         & & \textsc{Caltech101} \cite{fei2004learning}, \textsc{SUN09} \cite{choi2010exploiting} & & & & & SUN09, VOC2007\\
         \midrule
         
         \multirow{2}{*}{\textsc{PACS} \cite{li2017deeper}} & \multirow{2}{*}{2017} & \textsc{Caltech256} \cite{griffin2007caltech}, \textsc{Sketchy}\cite{Sketchy} & \multirow{2}{*}{9K} & \multirow{2}{*}{No} & \multirow{2}{*}{No} & \multirow{2}{*}{4} & \multirow{2}{*}{Photo, Art, Cartoon, Sketch}\\
         
         & & \textsc{TU-Berlin}\cite{TU-Berlin}, Google Images & & & & & \\
         \midrule
         
         \multirow{2}{*}{\textsc{WaterBird} \cite{sagawa2019distributionally}} & \multirow{2}{*}{2019} & \multirow{2}{*}{\textsc{CUB} \cite{wah2011caltech}, \textsc{Places} \cite{zhou2017places}} & \multirow{2}{*}{9K} & \multirow{2}{*}{No} & \multirow{2}{*}{No} & \multirow{2}{*}{2} & Correlations Between \\
         
         & & & & & & & Birds and Backgrounds\\
         \midrule
         
         \multirow{2}{*}{\textsc{DomainNet} \cite{peng2019moment}} & \multirow{2}{*}{2019} & \multirow{2}{*}{Internet} & \multirow{2}{*}{600K} & \multirow{2}{*}{No} & \multirow{2}{*}{No} & \multirow{2}{*}{6} & Sketch, Real, Quickdraw\\
         
         & & & & & & & Painting, Infograph, Clipart\\
         \midrule
         
         \textsc{Camelyon17} \cite{koh2021wilds} & 2021 & WILDS \cite{koh2021wilds} & 400K & No&No & 5 & Medical Centers\\
         \midrule
         
         \multirow{2}{*}{\textsc{FMoW} \cite{koh2021wilds}} & \multirow{2}{*}{2021} & \multirow{2}{*}{WILDS \cite{koh2021wilds}} & \multirow{2}{*}{500K} & \multirow{2}{*}{No} & \multirow{2}{*}{No} & 3 &  2002–2013, 2013–2016, 2016–2018\\
         & & & & & & 5 & Africa, Americas, Oceania, Asia, Europe\\
         \midrule
         
         \textsc{YFCC100M-FDG} \cite{fedora} & 2024 & \textsc{YFCC100M} \cite{thomee2016yfcc100m} & 90K & 1,570 & No & 3 & Before 1999, 2000-2009, 2010-2014\\
    \midrule
      \multicolumn{8}{c}{\textbf{Face Attribute and Distribution Shifts}} \\
      \midrule
         
         \multirow{2}{*}{\textsc{FairFace} \cite{fairface}} & \multirow{2}{*}{2019} & \multirow{2}{*}{Flickr, Twitter, Newspapers, Internet} & \multirow{2}{*}{108K} & 3 & \multirow{2}{*}{Yes} & \multirow{2}{*}{7} & Black, East Asian, Indian, Latino\\
         & & & & & & & Middle Eastern, Southeast Asian, White\\
         \midrule
         
         \textsc{UTK-FairFace} \cite{utk-fairface} & 2022 & \textsc{UTKFace} \cite{utk}, \textsc{FairFace} \cite{fairface} & 128K &3 & Yes & 2 & \textsc{UTKFace} \cite{utk}, \textsc{FairFace} \cite{fairface}\\
        \midrule

         \multirow{4}{*}{\textsc{\sysname{}}} & \multirow{4}{*}{2025} & \textsc{CelebA} \cite{CelebA}, \textsc{MetFaces} \cite{karras2020training} & \multirow{4}{*}{100K} & \multirow{4}{*}{39} & \multirow{4}{*}{Yes} & \multirow{4}{*}{4} & \multirow{4}{*}{Photo, Art, Cartoon, Sketch}\\

         & & \textsc{WikiART} \cite{artgan2018}, \textsc{Human-Art} \cite{ju2023human} & & & & & \\

         & & \textsc{IIIT-CFW} \cite{mishra2016iiit}, \textsc{iCartoonFace} \cite{zheng2020cartoon} & & & & & \\
         
         & & \textsc{CUFSF} \cite{zhang2011coupled}, \textsc{FS2K} \cite{fan2022facial}, Internet & & & & & \\
    \bottomrule
    \end{tabular}
    \label{tab:facedatasets}
\end{table*}

\section{More Details of Data Annotation}
\begin{figure}[h]
    \centering
    \includegraphics[width=0.8\linewidth]{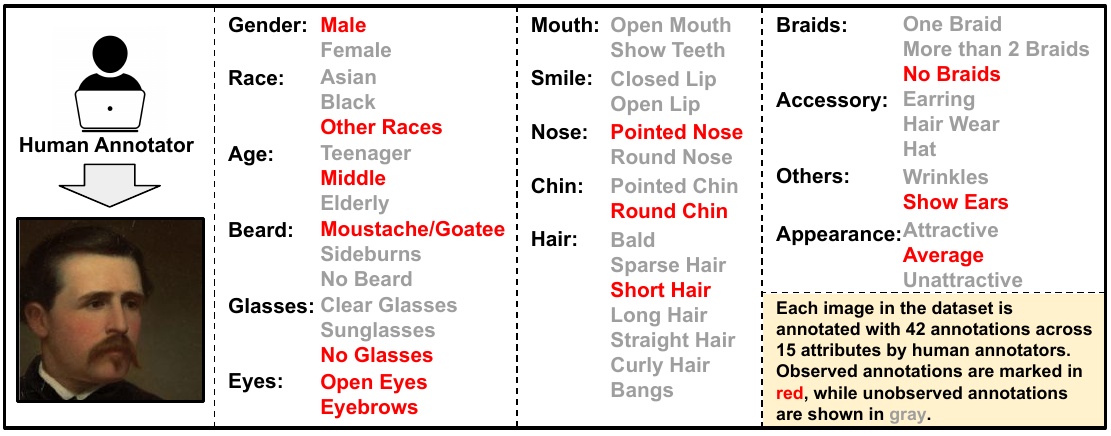}
    \caption{An example of data annotation. Each image in \textsc{\sysname{}} is annotated with 42 annotations across 15 attributes by human annotators.}
    \label{fig:annoexp}
\end{figure}

\begin{figure}[h]
    \centering
    \includegraphics[width=\linewidth]{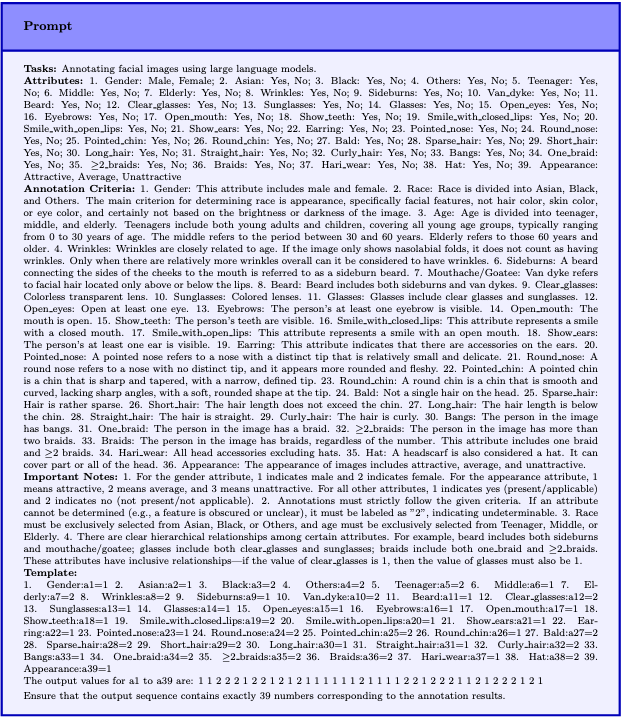}
    \caption{The prompt template used to generate 15 attribute annotations with LLMs.}
    \label{fig:prompt}
\end{figure}

\begin{itemize}[leftmargin=*]
    \item Figure~\ref{fig:annoexp} illustrates an example of data annotation for a sample image.
    \item The prompt template used to generate 15 attribute annotations with LLMs for input images sampled from different domains is shown in Figure \ref{fig:prompt}.
    \item During data annotation, we perform a comparative analysis between human annotations and those generated by large language models (LLMs), including GPT-4o, Claude 3.7 Sonnet, Gemini 2.5, and Kimi Latest. The comparison is conducted across 15 facial attributes using accuracy and F1-score, with human annotations serving as the ground truth.
Specifically, for single-label attributes such as gender, race, age, and appearance, accuracy is computed as:
\begin{align*}
    \text{Accuracy} = \frac{1}{N}\sum_{i=1}^N 1[y^{pred}_i=y^{true}_i]
\end{align*}
For the remaining multi-label attributes, accuracy is evaluated using the Jaccard accuracy \cite{murphy1996finley} (also known as the Jaccard Index or Intersection over Union):
\begin{align*}
    \text{Jaccard Accuracy}=\frac{1}{N}\sum_{i=1}^N \frac{|y^{pred}_i\cap y^{true}_i|}{|y^{pred}_i\cup y^{true}_i|}
\end{align*}
Here, the numerator represents the number of correctly predicted labels, while the denominator denotes the total number of unique labels from both the prediction and the ground truth.
\end{itemize}

\label{sec:app_dataannotation}

\section{Experiment Details}
\label{sec:app_implementationdetails}
All experiments are conducted on NVIDIA A100 and NVIDIA GeForce RTX 3090 GPUs, and the results are reported as the average performance over 3 runs.

\subsection{Feature Space Analysis}
\textbf{Evaluation Metrics.} 
We quantify the covariate shift across domains for all datasets using the Jensen-Shannon (JS) divergence \cite{endres2003new}, which is computed based on the Kullback-Leibler (KL) divergence as follows.
\begin{align*}
    D_{JS} = \frac{1}{2}\sum\nolimits_{i=1}^M\sum\nolimits_{j=1}^M \Big(\frac{1}{2}KL(D_i||\frac{D_i+D_j}{2}) + \frac{1}{2}KL(D_j||\frac{D_i+D_j}{2})\Big), \forall i\neq j
\end{align*}
where $D_i$ and $D_j$ are data features from any two different domains and $M$ denotes the number of domains.

\subsection{Fairness Learning}
\textbf{Evaluation Metrics.} 
We employ two widely used fairness evaluation metrics: the absolute difference of demographic difference ($\Delta$DP) \cite{dwork2012fairness} and that of equalized odds ($\Delta$EO) \cite{hardt2016equality}.
\begin{align*}
    &\Delta \text{DP} = |\mathbb{P}(\hat{Y}=1|Z=1)-\mathbb{P}(\hat{Y}=1|Z=-1)|\\
    &\Delta \text{EO} = \sum\nolimits_{y\in\{0,1\}}|\mathbb{P}(\hat{Y}=1|Z=1,Y=y) - \mathbb{P}(\hat{Y}=1|Z=-1,Y=y)|
\end{align*}
where $Y$ represents the ground-truth class label and $\hat{Y}$ denotes the predicted label.
A value of zero for these metrics indicates fair predictions, meaning the model's decisions are independent of the sensitive attribute $Z$.
Additionally, to analyze the stability of these baselines on a given dataset, we introduce two stability indicators, $\epsilon$ and $\pm$. $\epsilon$ quantifies the average absolute difference between any two baseline methods.
\begin{align*}
    \epsilon = \frac{1}{2}\sum\nolimits_{i=1}^N\sum\nolimits_{j=1}^N|\text{baseline}_i-\text{baseline}_j|, \forall i\neq j
\end{align*}
where $N$ denotes the number of baseline methods. 
The $\pm$ indicator represents the standard deviation of the performance across all baseline methods.  
Smaller values of both $\epsilon$ and $\pm$ indicate more stable and consistent performance across different baselines.

We conducted fairness learning experiments on the \textsc{CelebA}, \textsc{UTKFace}, \textsc{FairFace}, \textsc{UTK-FairFace}, and \textsc{\sysname{}} datasets, selecting age as the binary label. In \textsc{\sysname{}}, age is discretized into "elderly" and "non-elderly". For datasets where age is a continuous variable (\textit{e.g.,} \textsc{UTKFace} and \textsc{FairFace}), we binarize the labels by setting the threshold at 50 years old.
We then conducted experiments using gender and race as binary sensitive attributes, where race was standardized into "black" and "non-black". Note that since \textsc{CelebA} does not contain race information, we excluded it from experiments where race was the sensitive attribute.
Since this experiment does not involve domain partitioning, each dataset was randomly split into training and test sets in an 8:2 ratio. Each experiment was repeated three times, and the results were averaged. The fairness learning results, along with standard deviations, are reported in Table~\ref{tab:appendix_fairness_result}.

\begin{table*}[t!]
\centering
\scriptsize
\caption{Complete Fairness Learning Results Using Multiple Baselines. (Extended from Table \ref{tab:fairness_result} with Standard Deviations in Parentheses)}
\begin{threeparttable}
\begin{tabular}{c|c|c|c|c|c|c|c|c}
\toprule
 \textbf{Label} & \multirow{2.4}{*}{\textbf{Dataset}}& \multicolumn{7}{c}{\textbf{Accuracy}}\\

    \cmidrule(lr){1-1}
  \cmidrule(lr){3-9}
      \textbf{Sen.}&& LFR &  GSR & AD &
      CSAD &
      FNF & $\epsilon$ & $\pm$\\

    \midrule

 \multirow{5}{*}{\rotatebox{90}{\makecell[c]{\textbf{Age}\\\textbf{Gender}}}}&\textsc{Celeba}~\cite{CelebA}&78.92 (0.007) & 80.49 (0.004) & 78.48 (0.036) & 85.37 (0.009) & 85.96 (0.020) & 0.043 & 0.036\\

&\textsc{UTKFace}~\cite{utk}&78.13 (0.012) & 78.27 (0.003) & 78.93 (0.044) & 83.29 (0.005) & 83.50 (0.012) & 0.032 & 0.027 \\
 &\textsc{FairFace}~\cite{fairface}&87.32 (0.003) & 83.21 (0.009) & 83.87 (0.032) & 89.25 (0.003) & 89.69 (0.020) & 0.037 & 0.030 \\

 &\textsc{\utkfair{}}~\cite{utk-fairface}&86.43 (0.009) & 79.38 (0.007) & 78.40 (0.010) & 84.44 (0.011) & 83.87 (0.011) & 0.042 & 0.034 \\
 &\textsc{\sysname{}}&\textbf{75.06 (0.002)} & \textbf{74.50 (0.005)} & \textbf{75.90 (0.015)} & \textbf{78.49 (0.004)} & \textbf{79.28 (0.014)} & \textbf{0.026} & \textbf{0.021} \\

 \midrule

 \multirow{4}{*}{\rotatebox{90}{\makecell[c]{\textbf{Age}\\\textbf{Race}}}}&\textsc{UTKFace}~\cite{utk}&76.90 (0.018) & 76.37 (0.005) & 75.44 (0.010) & 78.81 (0.007) & 77.24 (0.003) & 0.015 & 0.012 \\

 &\textsc{FairFace}~\cite{fairface}&87.42 (0.005) & 84.23 (0.007) & 82.71 (0.007) & 88.90 (0.003) & 88.89 (0.004) & 0.034 & 0.028\\

 &\textsc{\utkfair{}}~\cite{utk-fairface}&84.67 (0.008) & 81.63 (0.009) & 81.62 (0.004) & 85.60 (0.004) & 85.19 (0.004) & 0.023 & 0.020 \\
 &\textsc{\sysname{}}&\textbf{73.53 (0.004)} & \textbf{73.97 (0.003)} &\textbf{ 74.70 (0.004)} & \textbf{76.37 (0.005) }& \textbf{75.30 (0.004)} & \textbf{0.014} & \textbf{0.011}\\

\bottomrule
\end{tabular}
\end{threeparttable}

\label{tab:appendix_fairness_result}

\end{table*}
\begin{table*}[t!]
\centering
\scriptsize
\vspace{-3mm}
\begin{threeparttable}
\begin{tabular}{c|c|c|c|c|c|c|c|c}
\toprule
 \textbf{Label} & \multirow{2.4}{*}{\textbf{Dataset}}& \multicolumn{7}{c}{\textbf{$\Delta\text{DP}$}} \\

    \cmidrule(lr){1-1}
  \cmidrule(lr){3-9}
      \textbf{Sen.}&& LFR &  GSR & AD &
      CSAD &
      FNF & $\epsilon$ & $\pm$\\

    \midrule

 \multirow{5}{*}{\rotatebox{90}{\makecell[c]{\textbf{Age}\\\textbf{Gender}}}}&\textsc{Celeba}~\cite{CelebA}& 0.062 (0.007) & 0.069 (0.002) & 0.067 (0.003) & 0.049 (0.002) & 0.047 (0.002) & 0.012 & 0.010 \\

&\textsc{UTKFace}~\cite{utk}& 0.021 (0.001) & 0.020 (0.004) & 0.021 (0.002) & 0.020 (0.001) & 0.019 (0.003) & \textbf{0.001} & \textbf{0.001}\\
 &\textsc{FairFace}~\cite{fairface} & 0.071 (0.003) & 0.025 (0.006) & 0.045 (0.007) & 0.025 (0.004) & 0.026 (0.005) & 0.023 & 0.020\\

 &\textsc{\utkfair{}}~\cite{utk-fairface}&0.045 (0.004) & 0.019 (0.001) & 0.021 (0.000) & 0.018 (0.000) & 0.018 (0.001) & 0.011 & 0.011\\
 &\textsc{\sysname{}}& \textbf{0.092 (0.001)} & \textbf{0.083 (0.003)} & \textbf{0.083 (0.002)} & \textbf{0.079 (0.002)} & \textbf{0.071 (0.002)} & 0.009 & 0.008 \\

 \midrule

 \multirow{4}{*}{\rotatebox{90}{\makecell[c]{\textbf{Age}\\\textbf{Race}}}}&\textsc{UTKFace}~\cite{utk}& 0.096 (0.004) & 0.121 (0.006) & 0.100 (0.003) & 0.083 (0.005) & 0.097 (0.003) & 0.016 & 0.014\\

 &\textsc{FairFace}~\cite{fairface}&0.144 (0.004) & 0.134 (0.005) & 0.129 (0.007) & 0.126 (0.004) & 0.123 (0.002) & 0.010 & 0.008 \\

 &\textsc{\utkfair{}}~\cite{utk-fairface}& 0.122 (0.007) & 0.094 (0.005) & 0.098 (0.005) & 0.092 (0.005) & 0.095 (0.005) & 0.013 & 0.013 \\
 &\textsc{\sysname{}} & \textbf{0.154 (0.006) }& \textbf{0.150 (0.006)} & \textbf{0.155 (0.005)} &\textbf{ 0.145 (0.004)} & \textbf{0.143 (0.006)} & \textbf{0.006} & \textbf{0.005}\\

\bottomrule
\end{tabular}
\end{threeparttable}

\end{table*}

\begin{table*}[t!]
\centering
\scriptsize

\vspace{-3mm}
\begin{threeparttable}
\begin{tabular}{c|c|c|c|c|c|c|c|c}
\toprule
 \textbf{Label} & \multirow{2.4}{*}{\textbf{Dataset}}&\multicolumn{7}{c}{\textbf{$\Delta\text{EO}$}}\\

    \cmidrule(lr){1-1}
  \cmidrule(lr){3-9}
      \textbf{Sen.}&& LFR &  GSR & AD &
      CSAD &
      FNF & $\epsilon$ & $\pm$ \\

    \midrule

 \multirow{5}{*}{\rotatebox{90}{\makecell[c]{\textbf{Age}\\\textbf{Gender}}}}&\textsc{Celeba}~\cite{CelebA}&0.062 (0.006) & 0.051 (0.015) & 0.051 (0.015) & 0.036 (0.002) & 0.036 (0.003) & 0.013 & 0.011\\

&\textsc{UTKFace}~\cite{utk}&0.063 (0.002) & 0.070 (0.009) & 0.070 (0.010) & 0.053 (0.002) & 0.052 (0.003) & 0.011 & 0.009\\
 &\textsc{FairFace}~\cite{fairface}&0.083 (0.002) & 0.064 (0.003) & 0.065 (0.004) & 0.059 (0.002) & 0.060 (0.001) & 0.010 & 0.010\\

 &\textsc{\utkfair{}}~\cite{utk-fairface}& 0.047 (0.002) & 0.040 (0.005) & 0.042 (0.004) & 0.033 (0.002) & 0.030 (0.002) & 0.009 & \textbf{0.007}\\
 &\textsc{\sysname{}}&\textbf{0.086 (0.005)} & \textbf{0.083 (0.004)} & \textbf{0.083 (0.004)} & \textbf{0.074 (0.002)} & \textbf{0.070 (0.002) }& \textbf{0.008} & \textbf{0.007}\\

 \midrule

 \multirow{4}{*}{\rotatebox{90}{\makecell[c]{\textbf{Age}\\\textbf{Race}}}}&\textsc{UTKFace}~\cite{utk}&0.098 (0.010) & 0.088 (0.001) & 0.089 (0.001) & 0.084 (0.001) & 0.087 (0.002) & \textbf{0.006} & \textbf{0.005}\\

 &\textsc{FairFace}~\cite{fairface}& 0.129 (0.003) & 0.120 (0.002) & 0.123 (0.007) & 0.113 (0.001) & 0.119 (0.001) & 0.007 & 0.006\\

 &\textsc{\utkfair{}}~\cite{utk-fairface}& 0.102 (0.008) & 0.086 (0.003) & 0.088 (0.001) & 0.086 (0.005) & 0.087 (0.002) & 0.007 & 0.007\\
 &\textsc{\sysname{}}&\textbf{0.142 (0.003)} &\textbf{ 0.136 (0.003)} & \textbf{0.142 (0.004)} & \textbf{0.136 (0.003) }& \textbf{0.132 (0.003)} & \textbf{0.006} & \textbf{0.005}\\

\bottomrule
\end{tabular}
\end{threeparttable}

\end{table*}

We evaluate the performance of all datasets in fairness learning using five widely adopted baseline methods: LFR~\cite{lfr}, GSR~\cite{gsr}, AD~\cite{ad}, CSAD~\cite{csad}, and FNF~\cite{fnf}. 
\begin{itemize}[leftmargin=*]
    \item \textbf{LFR} \cite{lfr} aligns distributions across protected groups to create fair representations while preserving data utility and demographic parity.
    \item \textbf{GSR} \cite{gsr} converts fairness constraints into classification error bounds via sample reweighting and margin adjustments.
    \item \textbf{AD} \cite{ad} removes bias through adversarial networks that suppress sensitive attribute correlations in features.
    \item \textbf{CSAD} \cite{csad} minimizes cross-sample mutual information to disentangle bias-invariant patterns from latent representations.
    \item \textbf{FNF} \cite{fnf} generates fair data via invertible flow models that transform biased latent distributions.

\end{itemize}

For methods like LFR and GSR that do not directly process image data, we extract features from images using a ResNet50 pre-trained on \textsc{ImageNet} and use these extracted features as model inputs. The implementations of LFR, GSR, and AD are based on the AIF360 repository\footnote{https://github.com/Trusted-AI/AIF360}, while CASD and FNF are implemented using their publicly available code.

\subsection{OOD Generalization}

We conducted domain generalization experiments on the \textsc{FairFace}, \textsc{UTK-FairFace}, and \textsc{\sysname{}} datasets, following the domain split settings of ~\cite{utk-fairface, fedora} for \textsc{FairFace} and \textsc{UTK-FairFace}. In this experiment, we used age and gender as binary classification targets. We adopted a leave-one-domain-out validation criterion, where for each domain in the dataset, we performed the following experiment three times: training on data from all other domains and testing on the held-out domain. The final result for each domain was obtained by averaging the three runs, and the overall experiment result was derived by averaging across all domains.

The performance of all datasets is evaluated using six widely recognized methods ERM~\cite{erm}, IRM~\cite{irm}, GDRO~\cite{gdro}, Mixup~\cite{mixup}, MMD~\cite{mmd}, MBDG~\cite{mbdg}.

\begin{itemize}[leftmargin=*]
    \item \textbf{ERM} \cite{erm} minimizes empirical training loss to achieve generalization.

    \item \textbf{IRM} \cite{irm} learns invariant predictors across environments by enforcing feature-to-label invariance in causal mechanisms.

    \item \textbf{GDRO} \cite{gdro} optimizes worst-case group performance via distributionally robust objectives and regularization for group shifts.

    \item \textbf{Mixup} \cite{mixup} augments data with linear interpolations of samples to encourage smoother decision boundaries and reduce spurious correlations.

    \item \textbf{MMD} \cite{mmd} aligns feature distributions across domains using maximum mean discrepancy minimization in kernel space.

    \item \textbf{MBDG} \cite{mbdg} synthesizes diverse training domains via generative models to improve adaptation to unseen target domains.

\end{itemize}

The MBDG method was implemented using its publicly available code, while other methods were implemented using the DomainBed repository \cite{gulrajani2020search}.

\subsection{OOD Detection}
\label{sec:app_ooddetection}

\begin{figure*}[t]
    \centering
    \includegraphics[width=\linewidth]{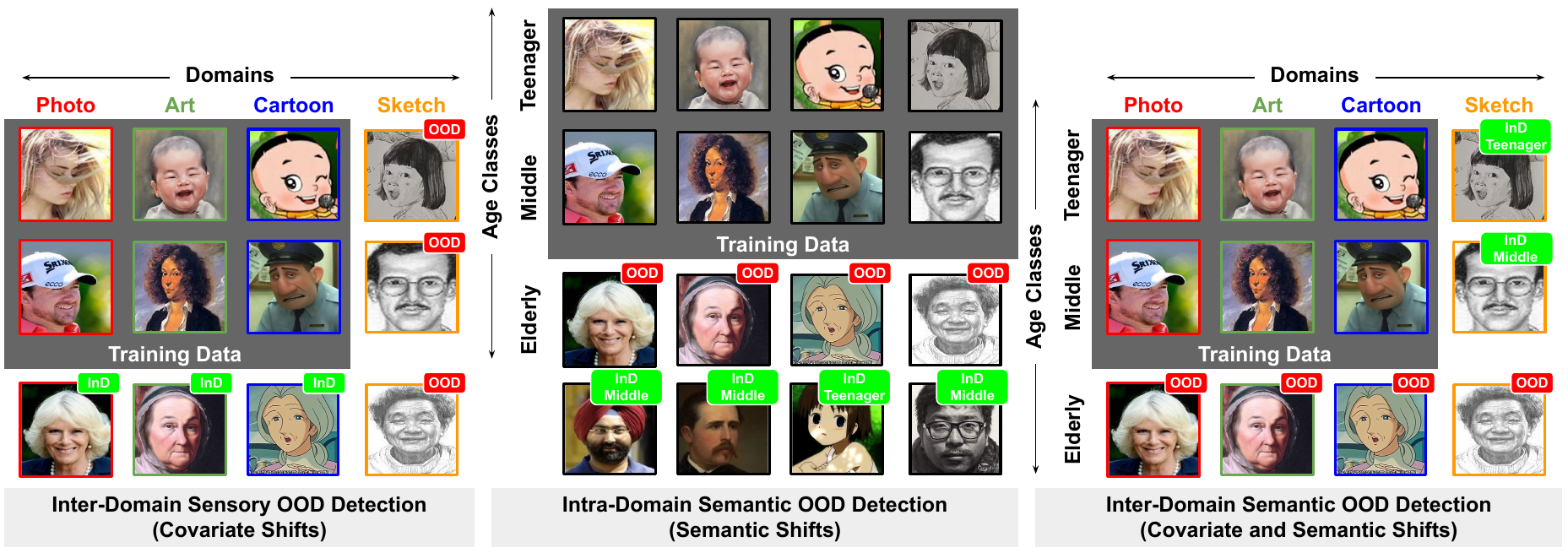}
    \caption{
    Experiment settings for OOD detection. 
    In inter-domain sensory OOD detection, test images exhibiting covariate shifts are considered OOD.
    In intra-domain and inter-domain semantic OOD detection, test images with semantic shifts are classified as OOD. 
    Specifically, in inter-domain semantic OOD detection, test images with only covariate shifts are considered as InD. 
    Note that both intra-domain and inter-domain semantic OOD detection experiments follow a $K+1$ classification setup, where the OOD detector's objective is to (1) distinguish between InD and OOD images and (2) classify InD images into one of the $K$ known classes.
    }
    \label{fig:ood_detection}
\end{figure*}

We conducted experiments on inter-domain sensory, intra-domain semantic, and inter-domain semantic OOD detection using the \textsc{FairFace}, \textsc{UTK-FairFace}, and \textsc{\sysname{}} datasets, following the same domain partitioning criteria as in OOD generalization. The experimental settings are illustrated in Figure~\ref{fig:ood_detection}. Specifically, we categorized age into three classes (teenager, middle, and elderly) and treated images of elderly individuals as semantic OOD samples.
For inter-domain sensory and inter-domain semantic OOD detection tasks, we adopted the leave-one-domain-out validation criteria, where images from the held-out domain were considered covariate OOD samples. In contrast, for intra-domain semantic OOD detection, we treated the entire dataset as a single domain.
For the OOD detection task, the training set consisted of 80\% of the data within the training data range shown in Figure~\ref{fig:ood_detection}, while the remaining 20\%, along with all other data, formed the test set.

For inter-domain sensory OOD detection and intra-domain semantic OOD detection, we use One-class SVM (ocSVM)~\cite{oc-svm}, DDU~\cite{ddu}, MSP~\cite{msp}, Energy~\cite{energy} and Entropy~\cite{entropy} as baseline methods. 
For inter-domain semantic OOD detection, we use EDst~\cite{edst}, SCONE~\cite{scone}, DAML~\cite{daml}, MEDIC~\cite{medic}, MADOD~\cite{maood} as baseline methods.

\begin{itemize}[leftmargin=*]
    \item \textbf{ocSVM} \cite{oc-svm} detects OOD samples by learning a hypersphere boundary around in-distribution data in a kernel-induced feature space.
    \item \textbf{DDU} \cite{ddu} estimates uncertainty via deterministic deep kernel Gaussian processes to distinguish in- and out-of-distribution samples
    \item \textbf{MSP} \cite{msp} uses maximum softmax probability as a confidence score, flagging low-confidence predictions as OOD.
    \item \textbf{Energy} \cite{energy} leverages energy-based scores from logits to detect OOD inputs without requiring distributional assumptions.
    \item \textbf{Entropy} \cite{entropy} identifies OOD samples by measuring prediction entropy, assuming higher uncertainty for anomalous inputs.
    \item \textbf{EDst} \cite{edst} employs domain-invariant feature alignment and simple statistical thresholds for open-domain generalization.
    \item \textbf{SCONE} \cite{scone} jointly optimizes OOD detection and generalization by leveraging unlabeled "wild" data for dual-purpose training.
    \item \textbf{DAML} \cite{daml} Enhances domain generalization via meta-learning with augmented domains to improve OOD detection robustness.
    \item \textbf{MEDIC} \cite{medic} learns dualistic meta-features to create generalizable decision boundaries for open-set domain shifts.
    \item \textbf{MADOD} \cite{maood} meta-trains models with invariance constraints to generalize OOD detection to unseen domains via g-invariance.

\end{itemize}

All ten OOD detection methods are implemented based on their original source code.

\subsection{Fairness-aware OOD Generalization}


We conducted fairness-aware OOD generalization experiments on the \textsc{FairFace}, \textsc{UTK-FairFace}, and \textsc{\sysname{}} datasets. Similar to the fairness experiments, we selected age as the binary label and chose gender and race as the binary sensitive attributes. The domain partitioning followed the same criteria as in the OOD generalization experiments, and we also adopted the leave-one-domain-out validation approach.

We evaluate all datasets using six recently published methods, FCR~\cite{utk-fairface}, EIIL~\cite{eiil}, FVAE~\cite{fvae}, FATDM~\cite{fatdm}, FLAIR~\cite{flair}, and FEDORA~\cite{fedora}.

\begin{itemize}[leftmargin=*]
    \item \textbf{FCR} \cite{utk-fairface} enforces fairness consistency across domains via distributionally robust regularization to maintain equity under covariate shifts.
    \item \textbf{EIIL} \cite{eiil} infers latent environments to learn invariant predictors that generalize fairly across unseen distributional shifts.
    \item \textbf{FVAE} \cite{fvae} disentangles sensitive attributes via contrastive distribution alignment in variational autoencoders for bias-free OOD generalization.
    \item \textbf{FLAIR} \cite{flair} learns fairness-aware domain-invariant representations by disentangling content/style factors and suppressing sensitive attributes while preserving utility, addressing both covariate and correlation shifts simultaneously.
    \item \textbf{FEDORA} \cite{fedora} addresses covariate shifts and correlation shifts simultaneously by learning fair invariant predictors through adversarial domain-invariant representation learning.
\end{itemize}

All five methods are implemented based on their original source code.



\section{Limitations and Discussion}
While \sysname{} offers a significant advancement in benchmarking fairness and robustness under visual domain shifts, several limitations remain. First, despite efforts to ensure diversity, the dataset's reliance on publicly available and web-crawled sources may introduce sampling biases or cultural imbalances that affect generalizability. Second, manual annotation, while thorough, may still suffer from subjectivity, particularly for attributes such as attractiveness, which are inherently ambiguous and context-dependent. Third, the current benchmark focuses primarily on facial imagery and may not generalize to broader fairness concerns in other modalities or domains. Future extensions could explore cross-modal fairness benchmarks, incorporate more fine-grained sensitive attributes, and develop automated tools to mitigate human annotation noise.
\label{sec:app_limitations}

\clearpage
\section*{NeurIPS Paper Checklist}

\begin{enumerate}

\item {\bf Claims}
    \item[] Question: Do the main claims made in the abstract and introduction accurately reflect the paper's contributions and scope?
    \item[] Answer: \answerYes{} 
    \item[] Justification: The abstract and introduction clearly present the motivation, contribution, and limitations of this study. 

\item {\bf Limitations}
    \item[] Question: Does the paper discuss the limitations of the work performed by the authors?
    \item[] Answer: \answerYes{} 
    \item[] Justification: The paper discusses the limitations in Appendix \ref{sec:app_limitations}.

\item {\bf Theory assumptions and proofs}
    \item[] Question: For each theoretical result, does the paper provide the full set of assumptions and a complete (and correct) proof?
    \item[] Answer: \answerNA{} 
    \item[] Justification: This paper is an experimental work without theoretical results. 

    \item {\bf Experimental result reproducibility}
    \item[] Question: Does the paper fully disclose all the information needed to reproduce the main experimental results of the paper to the extent that it affects the main claims and/or conclusions of the paper (regardless of whether the code and data are provided or not)?
    \item[] Answer: \answerYes{} 
    \item[] Justification: Detailed information required to reproduce the results is provided in the experimental sections and the Appendix. The code is attached, and we are committed to releasing it publicly upon acceptance.

\item {\bf Open access to data and code}
    \item[] Question: Does the paper provide open access to the data and code, with sufficient instructions to faithfully reproduce the main experimental results, as described in supplemental material?
    \item[] Answer: \answerYes{} 
    \item[] Justification: Our paper provides open access to the data and code, along with sufficient details for reproducing the main experimental results. 

\item {\bf Experimental setting/details}
    \item[] Question: Does the paper specify all the training and test details (e.g., data splits, hyperparameters, how they were chosen, type of optimizer, etc.) necessary to understand the results?
    \item[] Answer: \answerYes{} 
    \item[] Justification: See Experiments and Appendix. 

\item {\bf Experiment statistical significance}
    \item[] Question: Does the paper report error bars suitably and correctly defined or other appropriate information about the statistical significance of the experiments?
    \item[] Answer: \answerYes{} 
    \item[] Justification: All experiments are conducted multiple times, and standard deviations are provided in tables.

\item {\bf Experiments compute resources}
    \item[] Question: For each experiment, does the paper provide sufficient information on the computer resources (type of compute workers, memory, time of execution) needed to reproduce the experiments?
    \item[] Answer: \answerYes{} 
    \item[] Justification: Experiments compute resources are provided in Appendix \ref{sec:app_implementationdetails}.
    
\item {\bf Code of ethics}
    \item[] Question: Does the research conducted in the paper conform, in every respect, with the NeurIPS Code of Ethics \url{https://neurips.cc/public/EthicsGuidelines}?
    \item[] Answer: \answerYes{} 
    \item[] Justification: The research conducted in the paper conforms, in every respect, with the NeurIPS Code of Ethics. 

\item {\bf Broader impacts}
    \item[] Question: Does the paper discuss both potential positive societal impacts and negative societal impacts of the work performed?
    \item[] Answer: \answerNA{} 
    \item[] Justification: Only Technical reports 
    
\item {\bf Safeguards}
    \item[] Question: Does the paper describe safeguards that have been put in place for responsible release of data or models that have a high risk for misuse (e.g., pretrained language models, image generators, or scraped datasets)?
    \item[] Answer: \answerNA{} 
    \item[] Justification: There is no such misuse risk.

\item {\bf Licenses for existing assets}
    \item[] Question: Are the creators or original owners of assets (e.g., code, data, models), used in the paper, properly credited and are the license and terms of use explicitly mentioned and properly respected?
    \item[] Answer: \answerYes{} 
    \item[] Justification: All datasets and models used are cited. 

\item {\bf New assets}
    \item[] Question: Are new assets introduced in the paper well documented and is the documentation provided alongside the assets?
    \item[] Answer: \answerNA{} 
    \item[] Justification: The paper does not release new assets.

\item {\bf Crowdsourcing and research with human subjects}
    \item[] Question: For crowdsourcing experiments and research with human subjects, does the paper include the full text of instructions given to participants and screenshots, if applicable, as well as details about compensation (if any)? 
    \item[] Answer: \answerNA{} 
    \item[] Justification:  This paper does not involve crowdsourcing nor research with human subjects. 

\item {\bf Institutional review board (IRB) approvals or equivalent for research with human subjects}
    \item[] Question: Does the paper describe potential risks incurred by study participants, whether such risks were disclosed to the subjects, and whether Institutional Review Board (IRB) approvals (or an equivalent approval/review based on the requirements of your country or institution) were obtained?
    \item[] Answer: \answerNA{} 
    \item[] Justification:  This paper does not involve crowdsourcing nor research with human subjects. 

\item {\bf Declaration of LLM usage}
    \item[] Question: Does the paper describe the usage of LLMs if it is an important, original, or non-standard component of the core methods in this research? Note that if the LLM is used only for writing, editing, or formatting purposes and does not impact the core methodology, scientific rigorousness, or originality of the research, declaration is not required.
    \item[] Answer: \answerNA{} 
    \item[] Justification: The core method development in this research does not involve LLMs. 

\end{enumerate}

\end{document}